%% file: main.tex
\RequirePackage[svgnames]{xcolor}
\documentclass[11pt,letterpaper]{mystyle}
\usepackage[comma,authoryear]{natbib}
\usepackage{tocloft}
\input{preamble-common}

\usepackage{algorithm}
\usepackage{algorithmic}

\renewcommand{\TitleBoxLogo}{}
\title{Composable Soft-Prompt Distillation of RL Experts: Toward Continual Post-Training}
\runningtitle{Composable Soft-Prompt Distillation of RL Experts}
\expandafter\def\csname affnum@ucsb\endcsname{1}
\expandafter\def\csname affnum@cmu\endcsname{2}
\expandafter\def\csname affnum@ucsc\endcsname{3}
\expandafter\def\csname affnum@northwest\endcsname{4}
\expandafter\def\csname affnum@nvidia\endcsname{5}
\expandafter\def\csname affnum@uiuc\endcsname{6}
\expandafter\def\csname affnum@independent\endcsname{7}
\expandafter\def\csname affnum@ucla\endcsname{8}
\expandafter\def\csname affnum@semianalysis\endcsname{9}
\expandafter\def\csname affnum@sglang\endcsname{10}
\expandafter\def\csname affnum@michigan\endcsname{11}
\expandafter\def\csname affnum@princeton\endcsname{12}
\expandafter\def\csname affnum@lmsys\endcsname{13}
\expandafter\def\csname affnum@mindlab\endcsname{14}
\expandafter\def\csname affnum@columbia\endcsname{15}
\expandafter\def\csname affnum@apple\endcsname{16}
\expandafter\def\csname affnum@equal\endcsname{*}
\makeatletter
\newcommand{\paperauthor}[2]{\def\pa@nums{}\@for\pa@k:=#2\do{\edef\pa@nums{\ifx\pa@nums\@empty\else\pa@nums,\fi\csname affnum@\pa@k\endcsname}}\expandafter\author\expandafter[\pa@nums]{#1}}
\makeatother
\newcommand{\paperaffiliation}[2]{\affil[\csname affnum@#1\endcsname]{#2}}
\input{authors}
\input{affiliations}
\fancypagestyle{firststyle}{\fancyhf{}}
\begin{document}
\raggedbottom
\renewcommand{\topfraction}{0.9}
\renewcommand{\textfraction}{0.07}
\renewcommand{\floatpagefraction}{0.85}
\renewcommand{\dbltopfraction}{0.9}
\renewcommand{\dblfloatpagefraction}{0.85}
\setlength{\parindent}{0pt}
\linespread{1.05}\selectfont
\widowpenalty=10000 \clubpenalty=10000 \displaywidowpenalty=10000
\setlength{\parfillskip}{0pt plus 0.8\textwidth}
\begin{abstract}
\input{sections/abstract}
\end{abstract}
\maketitle
{\renewcommand{\thefootnote}{}\footnotetext{$^*$Equal contribution.}}
\begin{center}
\includegraphics[width=0.9\linewidth]{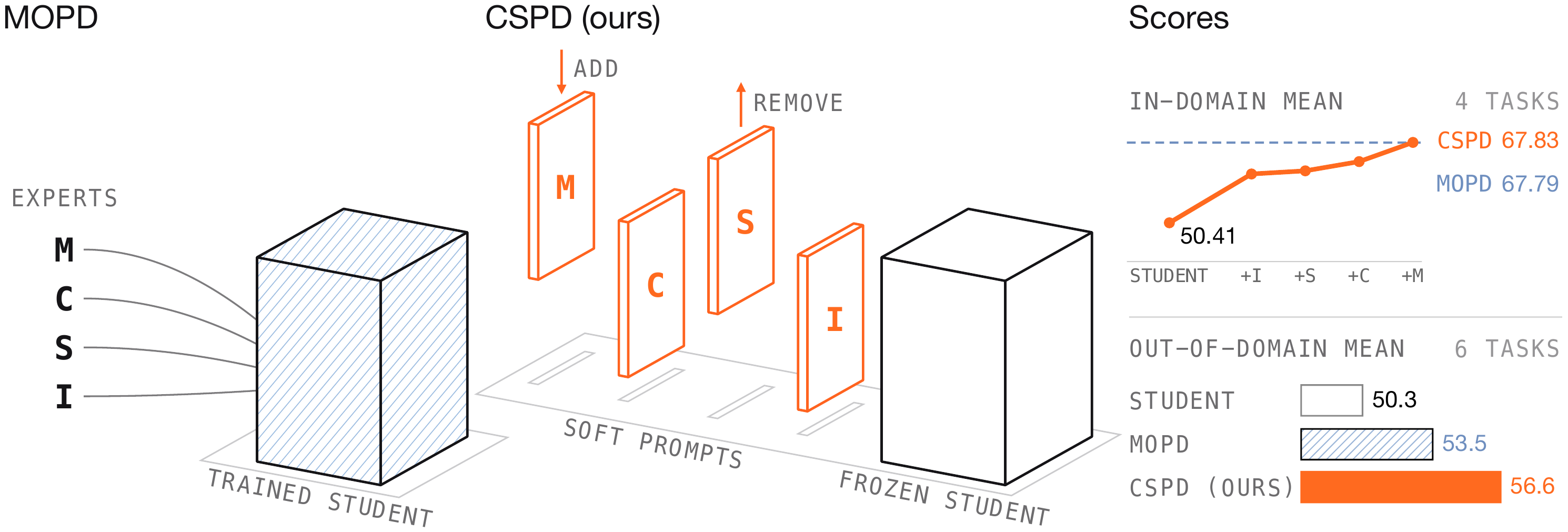}
\captionof{figure}{MOPD consolidates domain experts into the weights of a trained student. \cspd{} instead distills each expert into a separate soft prompt on the frozen student, and prompts are added, removed, and concatenated at inference. As experts are added one at a time, the in-domain mean of \cspd{} rises with every addition and ends on par with MOPD, while its out-of-domain mean is the highest.}\label{fig:teaser}
\end{center}
\input{body}

\bibliographystyle{plainnat}
\bibliography{references}
\clearpage
\appendix
\input{sections/appendix}

\end{document}

%% file: preamble-common.tex
\usepackage{colortbl}
\usepackage{array}
\usepackage{pgfplots}
\pgfplotsset{compat=1.18}
\usetikzlibrary{arrows.meta,positioning,fit}

\newcommand{\cspd}{\textsc{CSPD}}

%% file: authors.tex
\paperauthor{Zhen Zhang}{equal,ucsb} 
\paperauthor{Changyi Yang}{equal,cmu} 
\paperauthor{Xuehai He}{ucsc} 
\paperauthor{Chengzhi Liu}{ucsb} 
\paperauthor{Yutong Yin}{northwest} 
\paperauthor{Ziyue Liu}{ucsb} 
\paperauthor{Zhuoran Yin}{cmu} 
\paperauthor{Yangmin Li}{nvidia} 
\paperauthor{Zhihan Liu}{northwest} 
\paperauthor{Jisen Li}{uiuc} 
\paperauthor{Mingyu Jin}{independent} 
\paperauthor{Chuhan Li}{ucsb}
\paperauthor{Zhaotian Weng}{ucsb}
\paperauthor{Ruoyu He}{ucla} 
\paperauthor{Wenyao Gao}{semianalysis} 
\paperauthor{Jiaxin Deng}{sglang} 
\paperauthor{Chenwei Wu}{michigan}
\paperauthor{Hui Shen}{michigan} 
\paperauthor{Wenhao Chai}{princeton} 
\paperauthor{Chenyang Zhao}{lmsys} 
\paperauthor{Zhuofan Ying}{mindlab,columbia}
\paperauthor{Kieran Liu}{mindlab} %
\paperauthor{Andrew Chen}{mindlab} %
\paperauthor{Pony Ma}{mindlab} 
\paperauthor{Zhaoran Wang}{northwest} 
\paperauthor{Alkesh Patel}{apple} 
\paperauthor{Zhe Gan}{apple} 
\paperauthor{Xin Eric Wang}{ucsb} 

%% file: affiliations.tex
\paperaffiliation{ucsb}{University of California, Santa Barbara}
\paperaffiliation{cmu}{Carnegie Mellon University}
\paperaffiliation{ucsc}{University of California, Santa Cruz}
\paperaffiliation{northwest}{Northwestern University}
\paperaffiliation{nvidia}{NVIDIA}
\paperaffiliation{uiuc}{University of Illinois Urbana-Champaign}
\paperaffiliation{independent}{Independent Researcher}
\paperaffiliation{ucla}{University of California, Los Angeles}
\paperaffiliation{semianalysis}{SemiAnalysis}
\paperaffiliation{sglang}{SGLang-Omni}
\paperaffiliation{michigan}{University of Michigan}
\paperaffiliation{princeton}{Princeton University}
\paperaffiliation{lmsys}{LMSYS Org}
\paperaffiliation{mindlab}{Mind Lab}
\paperaffiliation{columbia}{Columbia University}
\paperaffiliation{apple}{Apple}

%% file: sections/abstract.tex
RL post-training typically produces domain experts that must then be consolidated into one model, and as new experts keep arriving, this consolidation becomes a continual process. Weight merging and multi-teacher on-policy distillation (MOPD) must re-consolidate the whole model whenever an expert is added, and can erode capabilities outside the target domains. We propose Composable Soft-Prompt Distillation (CSPD), which keeps the pre-RL model frozen, distills each RL expert independently into a soft prompt, and composes experts by concatenating their prompts without joint training. A single soft prompt suffices to distill an RL expert, recovering at least 90\% of the teacher's gain in 22 of 23 math, code, and agent settings. Independently trained prompts compose well. In a multi-expert setting, the average rises with every expert added, and the full composition matches MOPD (67.83 vs.\ 67.79). The composition generalizes best across six held-out benchmarks. Input-space adaptation adds little serving overhead, keeping throughput within 1.5\% of the unadapted backbone and reaching $2.40\times$ that of rank-4 LoRA. These results point toward continual post-training on a fixed backbone. Code is available at \url{https://github.com/UCSB-AI/CSPD}.

%% file: body.tex
\input{sections/introduction}
\input{sections/method}
\input{sections/experimental_setup}
\input{sections/single_expert}
\input{sections/composition}
\input{sections/incremental}
\input{sections/ood}
\input{sections/deployment}
\input{sections/interpretability}

\input{sections/discussion}
\input{sections/related_work}
\input{sections/conclusion}

%% file: sections/introduction.tex
\section{Introduction}
Reinforcement learning (RL) post-training is typically carried out one domain at a time, since each domain comes with its own data, verifier, and environment. A common recipe is therefore to train domain experts separately from a shared base model and then consolidate them into a single model \citep{m2rl2026,ma2026mopd}. Because new domains and stronger experts keep arriving, this consolidation is not a one-time step but a \textbf{continual process}. Ideally, each new expert would be absorbed like an installed module, such that (i) the model gains what the expert learned in its domain, (ii) everything the model could already do, in the other experts' domains and in general, is left intact, and (iii) the cost of the update depends only on the new expert, not on how many experts came before.

Existing approaches fall into two groups. The first consolidates all experts into a single set of weights. Joint multi-task RL must rerun RL on every domain whenever one is added. Sequential RL \citep{wang2026nemotroncascadescalingcascadedreinforcement} trains only on the new domain, but earlier gains are susceptible to irreversible damage from later stages. Weight merging \citep{ilharco2023arithmetic,yadav2023ties} combines experts by parameter arithmetic, where their updates can interfere. Multi-teacher on-policy distillation (MOPD) \citep{m2rl2026,ma2026mopd} must retrain the student against all teachers. All four also apply one fixed mixture to every request and overwrite the base weights, which can degrade performance outside the target domains \citep{wang2026opdgeneralization}. The second group keeps experts separate and selects them per request. Expert routing depends on a router that is hard to train and must be updated whenever an expert is added, cannot combine experts within a request, and must host a full model per expert. Merging LoRA experts per request \citep{huang2023lorahub,prabhakar2024loras} avoids these issues but adds adapter computation to every layer at every decoding step.

Both groups keep experts in model parameters. We instead make each expert part of the model's input, and propose \textbf{Composable Soft-Prompt Distillation (\cspd{})} (Figure~\ref{fig:method}). During training, \cspd{} keeps the pre-RL model frozen and distills each post-RL expert into its own soft prompt \citep{lester2021} of 256 vectors, using only that expert's verifier-filtered responses \citep{kim2016}. Each soft prompt is trained independently. To prepare the prompts for later combination, half of the training examples place a random context prefix before the prompt, so that it works regardless of what precedes it. At inference, the prompts are composed directly by concatenating them in the system message, without any joint training. Adding an expert therefore requires training only its prompt, removing an expert requires only dropping its prompt, and the backbone is never modified. Unlike merged or routed LoRA experts, the prompts are plug-in modules, with no merging coefficient or router to refit when the set of experts changes.

Merging experts in input space proves surprisingly effective. \textbf{A single soft prompt suffices to distill an RL expert.} Across 23 math, code, and agent settings, it recovers at least 90\% of the teacher's improvement over the pre-RL model in 22 of them, and matches or exceeds the teacher in 15. \textbf{Independently trained prompts exhibit strong composability.} In a multi-expert setting \citep{m2rl2026}, we distill soft prompts from experts in instruction following, science, code, and math, and add them to the frozen student one at a time, as in continual post-training. The four-domain average rises with every addition, from 50.41 to 67.83, and the full composition matches MOPD (67.79) without any joint training. \textbf{The composition generalizes best beyond the training domains.} Among the compared methods, it attains the highest mean score across six held-out benchmarks (56.6, versus 53.5 for MOPD). \textbf{A soft prompt learns behavior rather than words.} On a GPT-OSS math expert, the prompt closes 86.8\% of the teacher--student KL gap, mainly by making the student check its intermediate results, while written instructions with the same intent close at most 2.1\% (Section~\ref{sec:interp}).

\textbf{Input-space adaptation adds little serving overhead.} Each prompt only lengthens the context on a shared backbone, so requests with different prompts can be served efficiently in the same batch. On a single B200, soft-prompt throughput stays within 1.5\% of the unadapted backbone across seven concurrency levels. With 32 requests in flight, each using a distinct adapter, soft prompts deliver $2.40\times$ the output throughput of rank-4 LoRA. These results suggest a path toward continual post-training on a fixed backbone.

%% file: sections/method.tex
\section{Composable Soft-Prompt Distillation}\label{sec:method}

\begin{figure*}[t]
\centering
\includegraphics[width=\linewidth]{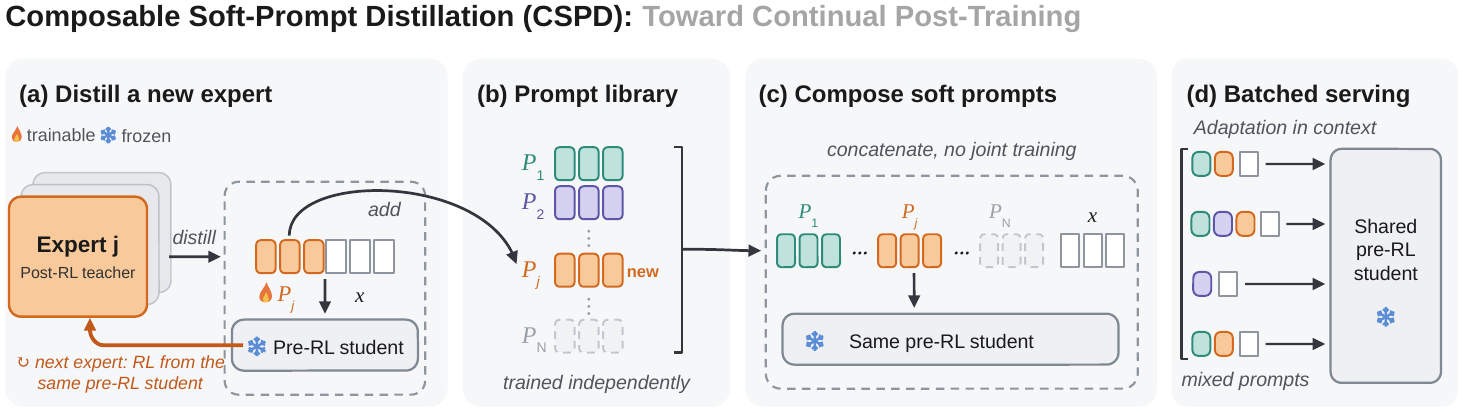}
\caption{Overview of composable soft-prompt distillation. Each RL expert is distilled into a separate prompt on the frozen pre-RL student (a). The resulting prompts form a library (b). Selected prompts are concatenated at inference (c), and requests with different prompts can share a batch on the same student (d).}
\label{fig:method}
\end{figure*}

In this section, we introduce Composable Soft-Prompt Distillation (\cspd{}), which distills the capabilities of post-RL experts into independently trained soft prompts on a shared frozen backbone. Figure~\ref{fig:method} illustrates the workflow.

\subsection{Distilling an RL Expert into a Soft Prompt}\label{sec:method-distill}
Let $p_{\theta_0}$ be the pre-RL model and $p_{\theta_j}$ an expert post-trained from it, with domain questions $\mathcal X_j$ and a verifier. We distill the expert into a soft prompt $P_j\in\mathbb R^{L\times d}$, a sequence of $L$ vectors in the backbone's $d$-dimensional embedding space. The template $\mathcal I(P_j,x)$ inserts $P_j$ into the system message, before the question $x$ (see Appendix~\ref{app:input-templates} for details). We use sequence-level distillation \citep{kim2016}. We sample teacher responses to questions in $\mathcal X_j$, keep those that pass the verifier, and train only $P_j$ on the resulting dataset $\mathcal D_j$ with a length-normalized negative log-likelihood over response tokens:
\begin{equation}\label{eq:prompt-distillation}
P_j^\star=\underset{P_j}{\arg\min}\;\mathbb E_{(x,y)\sim\mathcal D_j}\Big[-\frac{1}{|y|}\sum_{t=1}^{|y|}\log p_{\theta_0}\big(y_t\mid\mathcal I(P_j,x),y_{<t}\big)\Big].
\end{equation}
The backbone $\theta_0$ stays frozen, and each prompt is trained using only its own teacher and data.

\subsection{Composing Soft Prompts}\label{sec:method-compose}
To combine soft-prompt experts $1,\ldots,k$, we directly concatenate their soft tokens into $P_{1:k}=[P_1^\star;\ldots;P_k^\star]\in\mathbb R^{kL\times d}$ and insert it into the same template:
\begin{equation}\label{eq:compose}
p(y\mid x)=p_{\theta_0}\big(y\mid\mathcal I(P_{1:k},x)\big).
\end{equation}
No joint training is involved. In continual post-training, experts arrive over time. When a new expert arrives, we distill its prompt as above and add it to the existing ones, leaving the earlier prompts and the backbone unchanged. Removing an expert amounts to dropping its prompt.

\paragraph{Prefix-robust training for composition.}\label{sec:method-context}
A prompt is trained alone but follows other prompts at inference. To prepare it for this, we use prefix-robust training, which places a 128--256-token prefix, sampled from other questions in the same domain, before the soft prompt in half of the training examples. The prefix is excluded from the loss and is not used at inference. This makes composition more robust (Appendix~\ref{app:context-ablation}).

%% file: sections/experimental_setup.tex
\section{Experiments}\label{sec:experiments}
We first evaluate whether a soft prompt can recover a single RL expert (Section~\ref{sec:single}). We then evaluate the composition of independently distilled prompts in-domain (Section~\ref{sec:compose}) and in a continual post-training setting (Section~\ref{sec:incremental}), and evaluate out-of-domain generalization (Section~\ref{sec:ood}). Finally, we measure the cost of serving a distinct prompt per request (Section~\ref{sec:deploy}).

\subsection{Evaluation Metrics}\label{sec:setup}\label{sec:method-eval}
We report absolute task scores and the fraction of the RL teacher's performance gain recovered by the prompted student. Let $A_S$, $A_T$, and $A_{S+P}$ denote the student, teacher, and prompted-student scores, each either a single-task score or an average over multiple tasks. For $A_T>A_S$, recovery is
\begin{equation}\label{eq:recovery}
R=\frac{A_{S+P}-A_S}{A_T-A_S}\times100\%.
\end{equation}
A recovery of $100\%$ indicates that the prompted student matches the teacher's gain over the backbone. We report absolute scores alongside recovery because small teacher gains can produce large recovery ratios.

%% file: sections/single_expert.tex
\subsection{Distilling Individual RL Experts}\label{sec:single}
Before composing experts, we first test whether a single soft prompt can recover an individual RL expert, using public pre-RL/post-RL model pairs in math, code, and agent tasks. For each pair, we distill the RL teacher into a 256-vector prompt on its frozen pre-RL checkpoint and compare the prompted student (Student+SP) with both (Table~\ref{tab:single}). The table also gives model links and references, and evaluation details are in Appendices~\ref{app:inventory} and~\ref{app:agents}.

The 23 settings are deliberately diverse. They span math reasoning, code generation on LiveCodeBench, and multi-step agent tasks on SWE-bench Verified and Terminal-Bench 2.1. The models range from 7B to 35B parameters and include both dense and mixture-of-experts architectures from many model families, and their teachers were trained with different RL recipes and data. Across this range, the prompt recovers at least 90\% of the teacher's gain in 22 settings, and the prompted student matches or exceeds the teacher in 15 (Table~\ref{tab:single}). The trend holds for multi-step agent tasks as well, where the prompted student exceeds the teacher in five of six settings. For example, DeepSWE's student improves from 24.90\% to 45.65\% on SWE-bench Verified, compared with 44.05\% for its teacher. The lowest recovery, 88.08\%, is for MiMo-7B on code. Its RL training data are not publicly available, so we distill on AReaL code questions instead. This suggests that \cspd{} remains effective without the expert's original training data, although with some loss in recovery. Shorter prompts also recover much of the gain. On GPT-OSS math, 32 vectors reach 90.0\% recovery, below the 93.1\% of 256 vectors in the same sweep. Verifier filtering is also not strictly required. Training on responses of which only 69\% are correct still recovers 90.5\%, compared with 96.1\% for correct-only responses (Appendix~\ref{app:train-analysis}).

\begin{table}[htbp]
\caption{Distilling individual RL experts into soft prompts in math, code, and agent tasks. Student+SP is the best evaluated prompted-student score, and recovery follows Equation~\eqref{eq:recovery}. All values are percentages, and rows within each block are ordered by increasing recovery. \textit{Math:} unmarked rows use AIME24/25 + MATH500; $^{a}$AIME24 + AMC + MATH + Olympiad; $^{b}$AIME24/25 + HMMT25; $^{c}$AIME25/26 + MATH500. \textit{Code:} unmarked rows use LiveCodeBench v5/v6, and MiMo uses AReaL training data; $^{d}$Oct24--Feb25; $^{e}$test6 (1,055 questions); $^{f}$v3. \textit{Agent:} SWE-bench Verified uses 500 tasks, with avg@4 for Nemotron, DeepSWE, and Lego. Terminal-Bench 2.1 uses 89 tasks, avg@3. $^{\dagger}$The prompted student receives the teacher's predicted bug locations (Appendix~\ref{app:agents}).}\label{tab:single}\label{tab:math-full}\label{tab:code-full}\label{tab:agents}
\centering\scriptsize
\renewcommand{\arraystretch}{1.09}
\setlength{\tabcolsep}{2pt}
\begin{tabular*}{\linewidth}{@{}l@{\extracolsep{\fill}}llrrr>{\columncolor[HTML]{E6F1EF}}r@{}}
\toprule
\textbf{Pre-RL checkpoint} & \textbf{RL expert} & \textbf{Reference} & \textbf{Student} & \textbf{Teacher} & \textbf{Student+SP} & \textbf{Recovery (\%)}\\
\midrule
\multicolumn{7}{@{}l}{\textbf{\textit{(a) Math}}}\\
\midrule
\href{https://huggingface.co/XiaomiMiMo/MiMo-7B-SFT}{MiMo-7B-SFT} & \href{https://huggingface.co/XiaomiMiMo/MiMo-7B-RL}{MiMo-7B-RL} & \citealp{xiaomi2025mimounlockingreasoningpotential} & 63.77 & 73.18 & 72.25 & 90.15\\
\href{https://huggingface.co/tokyotech-llm/GPT-OSS-Swallow-20B-SFT-v0.1}{GPT-OSS-Swallow-20B-SFT} & \href{https://huggingface.co/tokyotech-llm/GPT-OSS-Swallow-20B-RL-v0.1}{GPT-OSS-Swallow-20B-RL} & \citealp{swallow2026gpt} & 76.53 & 90.21 & 89.67 & 96.06\\
\href{https://huggingface.co/tokyotech-llm/Qwen3-Swallow-8B-SFT-v0.2}{Qwen3-Swallow-8B-SFT} & \href{https://huggingface.co/tokyotech-llm/Qwen3-Swallow-8B-RL-v0.2}{Qwen3-Swallow-8B-RL} & \citealp{swallow2026qwen} & 63.68 & 83.01 & 82.28 & 96.22\\
\href{https://huggingface.co/allenai/Olmo-3-7B-Instruct-DPO}{OLMo-3-7B-Instruct-DPO} & \href{https://huggingface.co/allenai/Olmo-3-7B-Instruct}{OLMo-3-7B-Instruct} & \citealp{olmo2026olmo3} & 42.97 & 60.72 & 60.13 & 96.70\\
\href{https://huggingface.co/nvidia/Nemotron-Cascade-8B-Intermediate-ckpts/tree/main/Nemotron-Cascade-8B-SFT}{Nemotron-Cascade-8B-SFT} & \href{https://huggingface.co/nvidia/Nemotron-Cascade-8B-Intermediate-ckpts/tree/main/Nemotron-Cascade-8B-MathRL}{Nemotron-Cascade-8B-MathRL} & \citealp{wang2026nemotroncascadescalingcascadedreinforcement} & 85.25 & 89.34 & 89.30 & 99.18\\
\href{https://huggingface.co/tokyotech-llm/Qwen3-Swallow-30B-A3B-SFT-v0.2}{Qwen3-Swallow-30B-A3B-SFT} & \href{https://huggingface.co/tokyotech-llm/Qwen3-Swallow-30B-A3B-RL-v0.2}{Qwen3-Swallow-30B-A3B-RL} & \citealp{swallow2026qwen} & 72.91 & 86.25 & 86.23 & 99.79\\
\href{https://huggingface.co/PRIME-RL/Eurus-2-7B-SFT}{Eurus-2-7B-SFT} & \href{https://huggingface.co/PRIME-RL/Eurus-2-7B-PRIME}{Eurus-2-7B-PRIME}$^{a}$ & \citealp{cui2025processreinforcementimplicitrewards} & 36.06 & 49.63 & 49.91 & 102.07\\
\href{https://huggingface.co/PrimeIntellect/INTELLECT-MATH-SFT}{INTELLECT-MATH-SFT} & \href{https://huggingface.co/PrimeIntellect/INTELLECT-MATH}{INTELLECT-MATH}$^{a}$ & \citealp{prime2025math} & 45.47 & 54.11 & 54.62 & 105.89\\
\href{https://huggingface.co/LumiOpen/Llama-Poro-2-8B-Long-Math-Reasoning-SFT-Preview}{Poro-2-8B-Long-Math-SFT} & \href{https://huggingface.co/LumiOpen/Llama-Poro-2-8B-Long-Math-Reasoning-RL-Preview}{Poro-2-8B-Long-Math-RL}$^{c}$ & \citealp{hakala2026reasoning} & 68.34 & 72.43 & 73.88 & 135.47\\
\href{https://huggingface.co/Kwai-Klear/Klear-Reasoner-8B-SFT}{Klear-Reasoner-8B-SFT} & \href{https://huggingface.co/Kwai-Klear/Klear-Reasoner-8B}{Klear-Reasoner-8B}$^{b}$ & \citealp{su2026klearreasoneradvancingreasoningcapability} & 62.15 & 65.00 & 66.11 & 139.02\\
\midrule
\multicolumn{7}{@{}l}{\textbf{\textit{(b) Code (LiveCodeBench)}}}\\
\midrule
\href{https://huggingface.co/XiaomiMiMo/MiMo-7B-SFT}{MiMo-7B-SFT} & \href{https://huggingface.co/XiaomiMiMo/MiMo-7B-RL}{MiMo-7B-RL} & \citealp{xiaomi2025mimounlockingreasoningpotential} & 50.17 & 52.87 & 52.55 & 88.08\\
\href{https://huggingface.co/nvidia/Nemotron-Cascade-8B-Intermediate-ckpts/tree/main/Nemotron-Cascade-8B-SFT}{Nemotron-Cascade-8B-SFT} & \href{https://huggingface.co/nvidia/Nemotron-Cascade-8B-Intermediate-ckpts/tree/main/Nemotron-Cascade-8B-CodeRL}{Nemotron-Cascade-8B-CodeRL} & \citealp{wang2026nemotroncascadescalingcascadedreinforcement} & 55.73 & 69.39 & 69.56 & 101.21\\
\href{https://huggingface.co/Kwai-Klear/Klear-Reasoner-8B-SFT}{Klear-Reasoner-8B-SFT} & \href{https://huggingface.co/Kwai-Klear/Klear-Reasoner-8B}{Klear-Reasoner-8B} & \citealp{su2026klearreasoneradvancingreasoningcapability} & 53.44 & 57.27 & 57.34 & 101.83\\
\href{https://huggingface.co/Qwen/Qwen3-14B}{Qwen3-14B} & \href{https://huggingface.co/inclusionAI/AReaL-boba-2-14B-Open}{AReaL-boba-2-14B-Open}$^{d}$ & \citealp{fu2026areallargescaleasynchronousreinforcement} & 64.86 & 68.06 & 68.14 & 102.38\\
\href{https://huggingface.co/JetBrains/Mellum2-12B-A2.5B-Instruct-SFT}{Mellum2-12B-A2.5B-SFT} & \href{https://huggingface.co/JetBrains/Mellum2-12B-A2.5B-Instruct}{Mellum2-12B-A2.5B-Instruct}$^{e}$ & \citealp{kojic2026mellum2technicalreport} & 36.30 & 39.05 & 39.34 & 110.34\\
\href{https://huggingface.co/Qwen/Qwen3-8B}{Qwen3-8B} & \href{https://huggingface.co/inclusionAI/AReaL-boba-2-8B-Open}{AReaL-boba-2-8B-Open}$^{d}$ & \citealp{fu2026areallargescaleasynchronousreinforcement} & 58.54 & 62.20 & 62.96 & 120.83\\
\href{https://huggingface.co/allenai/Olmo-3-7B-Instruct-DPO}{OLMo-3-7B-Instruct-DPO} & \href{https://huggingface.co/allenai/Olmo-3-7B-Instruct}{OLMo-3-7B-Instruct}$^{f}$ & \citealp{olmo2026olmo3} & 28.50 & 34.10 & 36.54 & 143.44\\
\midrule
\multicolumn{7}{@{}l}{\textbf{\textit{(c) Agent: SWE-bench Verified}}}\\
\midrule
\href{https://huggingface.co/nvidia/Nemotron-Cascade-8B-Intermediate-ckpts/tree/main/Nemotron-Cascade-8B-SFT}{Nemotron-Cascade-8B-SFT} & \href{https://huggingface.co/nvidia/Nemotron-Cascade-8B}{Nemotron-Cascade-8B} & \citealp{wang2026nemotroncascadescalingcascadedreinforcement} &23.60&34.60&$33.75^{\dagger}$&$92.27^{\dagger}$\\
\href{https://huggingface.co/Qwen/Qwen3-32B}{Qwen3-32B} & \href{https://huggingface.co/agentica-org/DeepSWE-Preview}{DeepSWE-Preview} & \citealp{luo2025deepswe} &24.90&44.05&45.65&108.36\\
\href{https://huggingface.co/Qwen/Qwen3.5-35B-A3B}{Qwen3.5-35B-A3B} & \href{https://huggingface.co/Lego-X/qwen3_5_35b_a3b_ohsdk_200k_rl}{Lego-RL-35B-A3B} & \citealp{du2026legorlharnessnativereinforcementlearning} &64.00&70.40&72.25&128.91\\
\href{https://huggingface.co/NovaSky-AI/SWE-Gym-OpenHands-7B-Agent}{SWE-Gym-OpenHands-7B} & \href{https://huggingface.co/NovaSky-AI/SkyRL-Agent-7b-v0}{SkyRL-Agent-7b-v0} & \citealp{cao2025skyrl} &11.40&12.80&13.60&157.14\\
\midrule
\multicolumn{7}{@{}l}{\textbf{\textit{(d) Agent: Terminal-Bench 2.1}}}\\
\midrule
\href{https://huggingface.co/hamishivi/Qwen3.5-9B}{Qwen3.5-9B} & \href{https://huggingface.co/allenai/tmax-9b}{TMax-9B} & \citealp{ivison2026tmaxsimplerecipeterminal} &20.22&28.09&29.96&123.81\\
\href{https://huggingface.co/hamishivi/Qwen3.6-27B}{Qwen3.6-27B} & \href{https://huggingface.co/allenai/tmax-27b}{TMax-27B} & \citealp{ivison2026tmaxsimplerecipeterminal} &43.45&46.07&48.69&200.00\\
\bottomrule
\end{tabular*}
\end{table}

We next ask whether such prompts, trained independently for different domains, can be combined.

%% file: sections/composition.tex
\subsection{Composing Experts from Different Domains}\label{sec:compose}
We now further test whether soft prompts distilled independently from experts in different domains can be composed to combine their capabilities.

\paragraph{Experimental setting.}
We follow M2RL \citep{m2rl2026}, which first applies supervised fine-tuning to Qwen3-4B-Base \citep{qwen2025} and then trains a separate RL expert from this SFT model in each domain. We directly use their SFT model, M2RL-SFT, as the frozen student and their RL experts in four domains, math, code, science, and instruction following, as teachers. We do not retrain the SFT model or the RL experts, and we distill each soft prompt on the training data of the corresponding RL expert. These domains are evaluated on AIME24 \citep{aime}, LiveCodeBench v5 \citep{jain2025livecodebench}, GPQA-Diamond \citep{rein2024gpqa}, and IFBench \citep{pyatkin2025ifbench}, respectively. Evaluation uses temperature 0.6, a maximum generation length of 32k tokens, and sampling counts of $K=64/8/16/8$ across the four benchmarks.

Each expert is distilled independently into a 256-vector prompt with prefix-robust training (Section~\ref{sec:method-context}), using the Muon optimizer for 200 updates, and we use the final checkpoint. We evaluate the same four prompts in two configurations. \emph{Single-domain SP} attaches only the prompt of the domain being evaluated, and \cspd{} attaches all four prompts. We report two concatenation orders, MCSI (math, code, science, instruction following) and its reverse ISCM, and use MCSI by default. The input templates and training recipe are provided in Appendix~\ref{app:input-templates}.

\paragraph{Baselines.}
We compare against expert weight merging through Task Arithmetic (TA) \citep{ilharco2023arithmetic} and TIES \citep{yadav2023ties}, and against MOPD \citep{m2rl2026}. We also distill each expert into a LoRA adapter using the same data as the prompts and combine the adapters with TA or TIES. For TA, we report the best evaluated coefficient, which is 1.0 for the full-model RL experts and 0.5 for the LoRA adapters. Coefficient sweeps and baseline training settings are in Appendix~\ref{app:ta-sweep}.

\begin{table*}[t]
\caption{Single-domain and composition results. Full model, LoRA, and SP denote the adaptation representation. Scores and recovery are percentages. All rows use avg@$K$ with $K=64/8/16/8$ for AIME24/LCB/GPQA/IFBench.}\label{tab:compose}
\centering\small\setlength{\tabcolsep}{2pt}
\renewcommand{\arraystretch}{1.12}
\begin{tabular*}{\linewidth}{@{}l@{\extracolsep{\fill}}rrrrr>{\columncolor[HTML]{E6F1EF}}r@{}}
\toprule
\textbf{Method} & \textbf{AIME24} & \textbf{LCB v5} & \textbf{GPQA-D} & \textbf{IFBench} & \textbf{Avg.} & \textbf{Recovery (\%)}\\
\midrule
Student (no prompt) & 55.94 & 58.61 & 44.38 & 42.71 & 50.41 & 0.0\\
Individual domain experts (oracle routing) & 81.77 & 66.92 & 50.44 & 67.33 & 66.61 & 100.0\\
\midrule
\multicolumn{7}{@{}l}{\textit{Full model}}\\
\addlinespace[2pt]
Expert merge (TA, $\lambda=1.0$) & 78.85 & 64.52 & 54.14 & 71.58 & 67.27 & 104.1\\
Expert merge (TIES) & 81.25 & 65.42 & 57.07 & 66.12 & 67.47 & 105.2\\
MOPD & 83.28 & 66.24 & 55.21 & 66.42 & 67.79 & 107.2\\
\midrule
\multicolumn{7}{@{}l}{\textit{LoRA}}\\
\addlinespace[2pt]
LoRA distill + TA ($\lambda=0.5$) & 80.21 & 63.10 & 56.22 & 62.46 & 65.50 & 93.1\\
LoRA distill + TIES & 81.41 & 64.00 & 56.66 & 62.67 & 66.19 & 97.3\\
\midrule
\multicolumn{7}{@{}l}{\textit{Soft prompt}}\\
\addlinespace[2pt]
Single-domain SP & 80.89 & 64.67 & 49.24 & 66.71 & 65.38 & 92.4\\
\cspd{} w/o prefix-robust training & 76.98 & 64.15 & 51.89 & 65.00 & 64.50 & 87.0\\
\cspd{} (ISCM) & 79.90 & 65.72 & 57.92 & 67.33 & 67.72 & 106.8\\
\textbf{\cspd{} (MCSI)} & 80.47 & 64.67 & 57.77 & 68.42 & \textbf{67.83} & \textbf{107.5}\\
\bottomrule
\end{tabular*}
\end{table*}

\paragraph{Concatenated prompts recover the combined expert gain.}
Table~\ref{tab:compose} reports the single-domain and composition results. Composing the four prompts raises the student's average score from 50.41 to 67.83, recovering approximately 107.5\% of the aggregate expert gain and exceeding the average of the individual experts (66.61), which corresponds to an oracle router that sends each benchmark to its own domain expert. The same prompts evaluated individually average 65.38. Composition raises GPQA-Diamond from 49.24 to 57.77 and IFBench from 66.71 to 68.42, while AIME24 decreases slightly from 80.89 to 80.47. The reverse order ISCM gives a similar average of 67.72. Independently learned modules can therefore contribute beyond their own training domains, although composition does not improve every domain uniformly.

\paragraph{Training with varied context helps composition.}
Each prompt is trained alone but used after other prompts at inference. Prefix-robust training (Section~\ref{sec:method-context}) exposes it to varied preceding text during training. This makes the composition more stable (Table~\ref{tab:compose} and Appendix~\ref{app:context-ablation}).

\paragraph{Input-space composition matches weight-space consolidation.}
\cspd{} achieves an average score on par with MOPD (67.79) and higher than expert TIES (67.47) and expert TA (67.27). It also exceeds LoRA distillation followed by TA (65.50) or TIES (66.19), where the LoRA adapters use the same distillation data as the soft prompts. Performance varies by domain. \cspd{} scores higher than MOPD on GPQA-Diamond and IFBench, while MOPD performs better on AIME24 and LiveCodeBench. Thus, composing in input space reaches the in-domain performance of consolidating in weight space, while every expert remains a separate module that was trained without access to the others.

%% file: sections/incremental.tex
\subsection{Adding and Removing Experts}\label{sec:incremental}
We next evaluate adding experts one at a time, using the four prompts from Section~\ref{sec:compose}, keeping the backbone and all prompts fixed throughout.

\begin{figure*}[htbp]
\centering
\includegraphics[width=\textwidth]{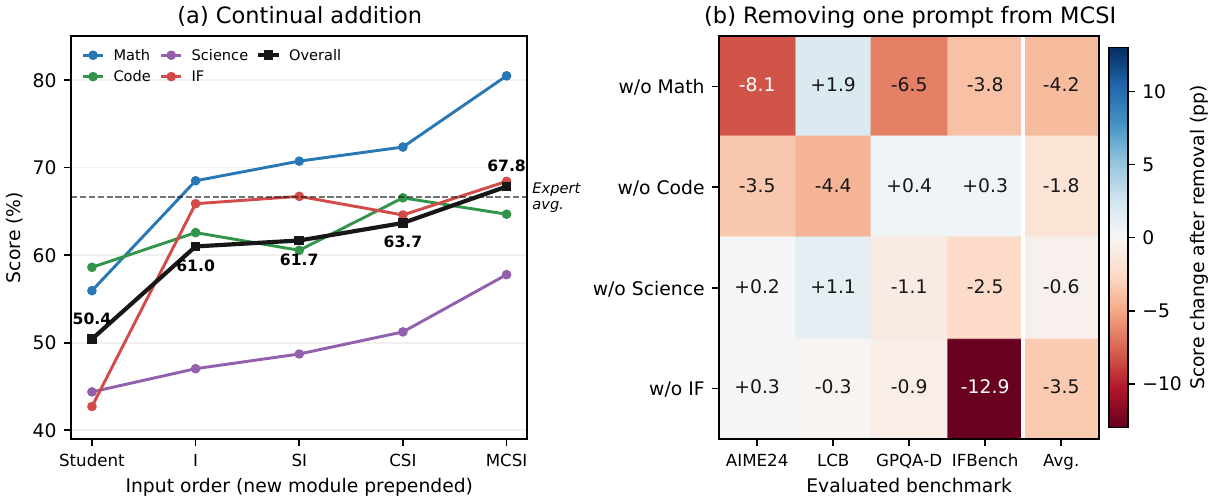}
\caption{Capability expansion and module interactions. (a) Independently trained prompts are prepended in arrival order I, S, C, M, starting from the unprompted student. The dashed line marks the average of the individual experts. Curves show each domain and the overall average. (b) Score change on each benchmark (columns) after removing one prompt (rows) from MCSI, preserving the remaining order. Negative values mean the removed prompt was helping. Both panels use the four prompts of Table~\ref{tab:compose}.}\label{fig:incremental}\label{fig:loo}
\end{figure*}

\paragraph{Continual addition.}
Starting from the unprompted student, experts arrive in the order instruction following (I), science (S), code (C), and math (M). Each new prompt is placed before the existing ones, yielding the inputs
\[
\mathrm{Student}\rightarrow\mathrm{I}\rightarrow\mathrm{SI}\rightarrow\mathrm{CSI}\rightarrow\mathrm{MCSI},
\]
where the leftmost module is the most recent. The final input is the default composition of Table~\ref{tab:compose}. We evaluate all four domains after each addition. The average rises from the student's 50.41 to 60.99, 61.67, 63.67, and 67.83 (Figure~\ref{fig:incremental}a), ending above the average of the individual experts (66.61), without retraining any existing module.

Individual domains exhibit both positive and negative interactions. Adding code reduces IFBench from 66.71 to 64.58, while subsequently adding math raises it to 68.42. Appending each new prompt instead gives a similar trajectory (Appendix~\ref{app:prepend}).

\paragraph{Module removal and cross-domain contributions.}
To examine each module's contribution, we remove one prompt at a time from the MCSI composition while preserving the relative order of the remaining prompts (Figure~\ref{fig:loo}b). Each row removes one prompt and each column is a benchmark. We report the score after removal minus the full-composition score, so a negative value means the removed prompt was helping.

In this configuration, every module contributes positively to its own domain and to the four-domain average. Removing math causes the largest average drop, at 4.16 points, including a 6.54-point decrease on GPQA-Diamond. Removing instruction following reduces IFBench by 12.88 points. Contributions also include negative interactions, as removing math improves LiveCodeBench by 1.87 points. The reverse order ISCM shows the same pattern (Appendix~\ref{app:loo-iscm}).

Experts therefore interact: some transfer across domains, as math does to science, while others interfere, as code briefly does with instruction following. Individual domains can dip as experts are added, but the overall average rises with every addition.

\begin{figure*}[!htbp]
\centering
\includegraphics[width=\textwidth]{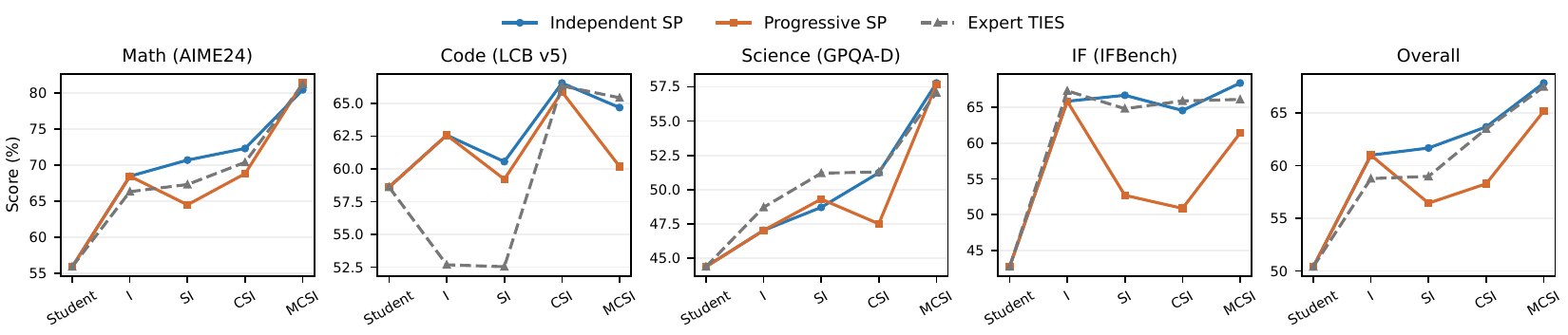}
\caption{Independent SP training, progressive SP training, and expert TIES merging by domain and overall. Experts arrive as I, S, C, M, and each new SP is prepended. All curves start at the unprompted student. The two SP strategies share the initial I prompt. TIES merges the same domain subset at each stage (the first stage uses the IF expert).}\label{fig:progressive-ablation}
\end{figure*}

\paragraph{Comparison with progressive training and merging.}
A natural alternative is to train each new prompt in the context of the earlier ones, as in Progressive Prompts \citep{razdaibiedina2023progressive}, where each new prompt is trained with all previous prompts frozen in front of it, so it can adapt to the existing composition. Because progressive training places each new prompt in front of the previous ones, we compare it with independent prompts in the same prepended order (I, SI, CSI, MCSI), with the same optimizer and number of updates. Both start from the same I prompt. Independent training gives a higher average at every stage (Figure~\ref{fig:progressive-ablation}): 61.67 versus 56.44 after S, 63.67 versus 58.28 after C, and 67.83 versus 65.17 after M. Progressive training also erodes the earliest capability, reducing IFBench to 52.71, 50.88, and 61.42 at these stages. Fitting a new module to a particular composition therefore does not help later joint use, and freezing earlier prompts alone does not preserve their behavior. Independent training is therefore both simpler and more effective in this setting. Figure~\ref{fig:progressive-ablation} also shows expert TIES merging over the same domain subsets. Independent prompts score higher than TIES on average at every stage, ending at 67.83 versus 67.47. The IF expert alone lowers LiveCodeBench to 52.69, well below the student's 58.61, and merging it with the science expert keeps it at 52.54. Updating the weights for one domain thus degrades another, a form of catastrophic forgetting. Independent prompts leave the backbone unchanged and keep LiveCodeBench at 62.57 and 60.55 at these stages.

%% file: sections/ood.tex
\subsection{Out-of-Domain Generalization}\label{sec:ood}
Consolidating experts into the weights can approach the experts in-domain while losing capabilities elsewhere \citep{wang2026opdgeneralization}. We therefore evaluate the MCSI composition of all four prompts from Table~\ref{tab:compose} on six benchmarks outside its four training domains.

We compare the composed prompts with the unprompted student, MOPD, expert TIES merging, and LoRA distillation with TA on six benchmarks (Table~\ref{tab:ood}, setting in Appendix~\ref{app:ood-protocol}). They cover refusal behavior on XSTest \citep{rottger2024xstest}, truthfulness on TruthfulQA \citep{lin2022truthfulqa}, tool calling on BFCL \citep{patil2025bfcl}, closed-book knowledge on TriviaQA \citep{joshi2017triviaqa}, difficult reasoning tasks on BBEH \citep{kazemi2025bbeh}, and financial reasoning on FinQA \citep{chen2021finqa}.

\begin{table}[htbp]
\caption{Out-of-domain scores (\%). \cspd{} uses the MCSI composition from Table~\ref{tab:compose}. The mean weights each benchmark equally and is computed from the displayed scores. Bold marks each column maximum. Parentheses give changes from the student.}\label{tab:ood}
\centering\small\renewcommand{\arraystretch}{1.16}
\setlength{\tabcolsep}{2pt}
\begin{tabular*}{\linewidth}{@{}l@{\extracolsep{\fill}}rrrrrrr@{}}
\toprule
\textbf{Method} & \textbf{XSTest} & \textbf{TruthfulQA} & \textbf{BFCL} & \textbf{TriviaQA} & \textbf{BBEH} & \textbf{FinQA} & \textbf{Mean}\\
\midrule
Student (no prompt) & 80.4 & \textbf{64.6} & 47.6 & 31.5 & 11.1 & 66.3 & 50.3\\
\midrule
\multicolumn{8}{@{}l}{\textit{Full model}}\\
Expert merge (TIES) & 82.4\,\textcolor[HTML]{176849}{\scriptsize $(+2.0)$} & 63.7\,\textcolor[HTML]{A23D32}{\scriptsize $(-0.9)$} & 48.5\,\textcolor[HTML]{176849}{\scriptsize $(+0.9)$} & \textbf{33.6}\,\textcolor[HTML]{176849}{\scriptsize $(+2.1)$} & 25.6\,\textcolor[HTML]{176849}{\scriptsize $(+14.5)$} & 70.9\,\textcolor[HTML]{176849}{\scriptsize $(+4.6)$} & 54.1\,\textcolor[HTML]{176849}{\scriptsize $(+3.9)$}\\
MOPD & 84.2\,\textcolor[HTML]{176849}{\scriptsize $(+3.8)$} & 61.0\,\textcolor[HTML]{A23D32}{\scriptsize $(-3.6)$} & 49.4\,\textcolor[HTML]{176849}{\scriptsize $(+1.8)$} & 31.9\,\textcolor[HTML]{176849}{\scriptsize $(+0.4)$} & 23.9\,\textcolor[HTML]{176849}{\scriptsize $(+12.8)$} & 70.8\,\textcolor[HTML]{176849}{\scriptsize $(+4.5)$} & 53.5\,\textcolor[HTML]{176849}{\scriptsize $(+3.3)$}\\
\midrule
\multicolumn{8}{@{}l}{\textit{LoRA}}\\
LoRA distill + TA ($\lambda=0.5$) & 85.1\,\textcolor[HTML]{176849}{\scriptsize $(+4.7)$} & 63.8\,\textcolor[HTML]{A23D32}{\scriptsize $(-0.8)$} & 47.5\,\textcolor[HTML]{A23D32}{\scriptsize $(-0.1)$} & 32.7\,\textcolor[HTML]{176849}{\scriptsize $(+1.2)$} & 23.0\,\textcolor[HTML]{176849}{\scriptsize $(+11.9)$} & 70.9\,\textcolor[HTML]{176849}{\scriptsize $(+4.6)$} & 53.8\,\textcolor[HTML]{176849}{\scriptsize $(+3.6)$}\\
\midrule
\multicolumn{8}{@{}l}{\textit{Soft prompt}}\\
\cspd{} (MCSI) & \textbf{88.2}\,\textcolor[HTML]{176849}{\scriptsize $(+7.8)$} & 64.2\,\textcolor[HTML]{A23D32}{\scriptsize $(-0.4)$} & \textbf{55.6}\,\textcolor[HTML]{176849}{\scriptsize $(+8.0)$} & 33.5\,\textcolor[HTML]{176849}{\scriptsize $(+2.0)$} & \textbf{25.8}\,\textcolor[HTML]{176849}{\scriptsize $(+14.7)$} & \textbf{72.5}\,\textcolor[HTML]{176849}{\scriptsize $(+6.2)$} & \textbf{56.6}\,\textcolor[HTML]{176849}{\scriptsize $(+6.4)$}\\
\bottomrule
\end{tabular*}
\end{table}

\cspd{} achieves the highest mean score (56.6) and leads on four of the six benchmarks. Its largest margin over the strongest comparison on each benchmark is on BFCL, where it scores 55.6 versus 49.4 for MOPD. It also leads on XSTest by 3.1 points over LoRA and on FinQA by 1.6 points over both expert TIES and LoRA. On BBEH, the gain over the student is large (25.8 versus 11.1), although expert merging achieves a similar score (25.6). The student retains the highest TruthfulQA score (64.6 versus 64.2 for \cspd{}), and expert merging leads on TriviaQA by 0.1 point. Overall, \cspd{} improves over the student on five of the six benchmarks and changes the sixth by only 0.4 points, whereas MOPD loses 3.6 points on TruthfulQA. Adding capabilities through the input, with the backbone frozen, thus largely preserves the student's general capabilities.

%% file: sections/deployment.tex
\subsection{Serving Efficiency}\label{sec:deploy}
A soft prompt adds parameters only at the input. It is processed once during prefill, and every later decoding step only attends to its cached keys and values, so relative to the backbone the only cost is a slightly longer context (Appendix~\ref{app:serving}). A LoRA adapter, by contrast, adds computation to every layer at every decoding step. Because every expert is an input to the same frozen model, requests that use different experts, or different combinations of them, can also share one batch. We measure the cost of this by assigning a distinct adapter to every request, the extreme case of per-user personalization, and comparing throughput and first-token latency against the same backbone served without adapters.

\paragraph{Experimental setting.}
We serve GPT-OSS-20B in BF16 on a single NVIDIA B200 with SGLang. We compare the unadapted backbone, 256-vector soft prompts, and unmerged rank-4 LoRA adapters served with the CSGMV backend. Each request has 20,480 text input tokens and generates 512 tokens, simulating an agent setting. SP adds 256 vectors to the input. We use matched text inputs across configurations and register the synthetic adapters before timing generation.

We vary the concurrency limit $C$ over $1,2,4,8,16,24,32$. After a warm-up burst of $C$ requests, a closed-loop client issues $8C$ measured requests, starting a new one whenever a slot becomes available. All three configurations use the same KV-cache budget at each concurrency level, with 524,288 tokens through $C=16$ and 1,048,576 at $C=24$ and $32$. We report aggregate output tokens per second from the first request submission to the last completion, and the 95th percentile of the time from request submission to receipt of the first token.

\begin{figure*}[htbp]
\centering
\includegraphics[width=\linewidth]{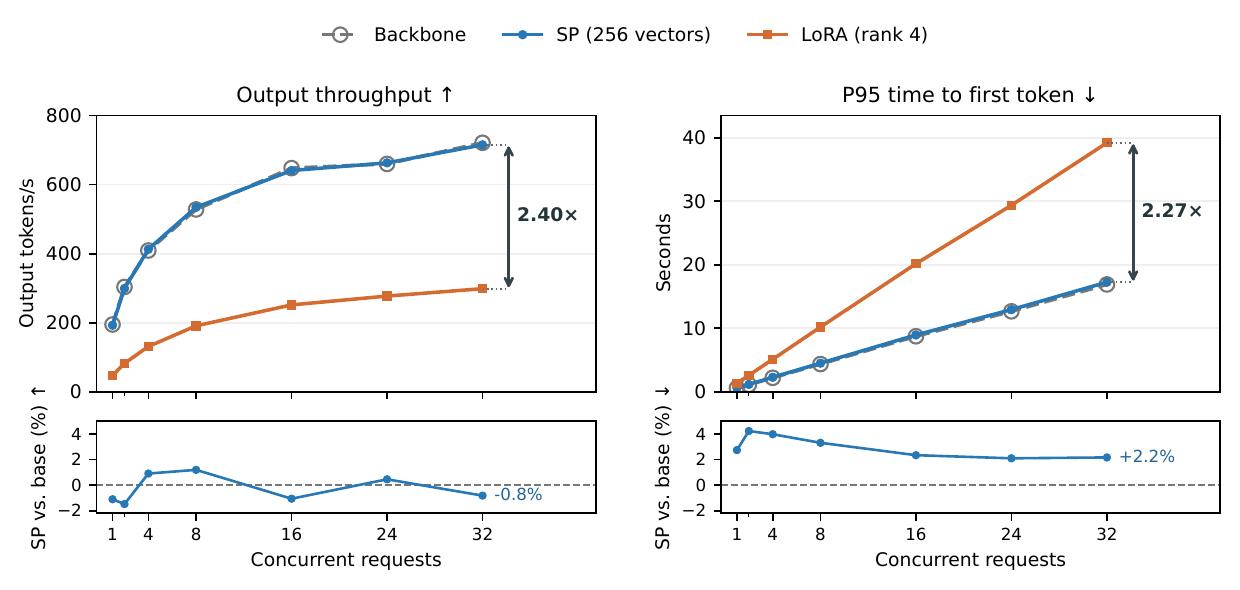}
\caption{Request-specific adapter serving on one B200. Higher throughput (left) and lower P95 time to first token (right) are better. At 32 concurrent requests, the arrows show the SP-to-LoRA throughput ratio and the LoRA-to-SP latency ratio. The lower panels show SP's percentage change relative to the unadapted backbone, with zero indicating equal performance.}\label{fig:unique-adapter-serving}
\end{figure*}

\paragraph{Multi-adapter serving.}
SP throughput stays within 1.5\% of the unadapted backbone across the seven concurrency levels (Figure~\ref{fig:unique-adapter-serving}). Relative to LoRA, the advantage is largest with one request in flight, where SP delivers $4.12\times$ the output throughput of LoRA. At 32 concurrent requests, SP serves 715.4 output tokens/s, compared with 721.3 for the backbone and 298.5 for LoRA. At this operating point, SP delivers $2.40\times$ the throughput of LoRA and reduces P95 time to first token from 39.21 to 17.26 seconds, a $2.27\times$ reduction, close to the backbone's 16.89 seconds.

\paragraph{Speculative decoding.}
Soft prompts also remain compatible with draft models trained on the backbone. Adding a 256-vector prompt reduces the average acceptance length from 2.34 to 2.28 for EAGLE-3 \citep{li2025eagle3} ($-2.6\%$) and from 2.96 to 2.82 for DFlash \citep{chen2026dflash} ($-4.7\%$). Acceptance length does not directly measure end-to-end latency.

%% file: sections/interpretability.tex
\section{What Does a Soft Prompt Learn?}\label{sec:interp}
We analyze one expert from Table~\ref{tab:single}, the 256-vector soft prompt that distills GPT-OSS-Swallow-20B-RL into GPT-OSS-Swallow-20B-SFT for math. The teacher, the student, and the prompted student answer 100 held-out AIME and MATH-500 questions. Each model then reads the same answers, and at every response position we compare its next-token distribution with the teacher's. See Appendix~\ref{app:interpretability} for details.

\textbf{The nearest word of a soft prompt vector does not tell what the vector learned.} Across six models, 84--100\% of the vectors keep their initialization token as the nearest word, even though in five of them the vectors move far from that token (median cosine similarity 0.25--0.34).

\begin{figure*}[!htbp]
\centering
\includegraphics[width=\textwidth]{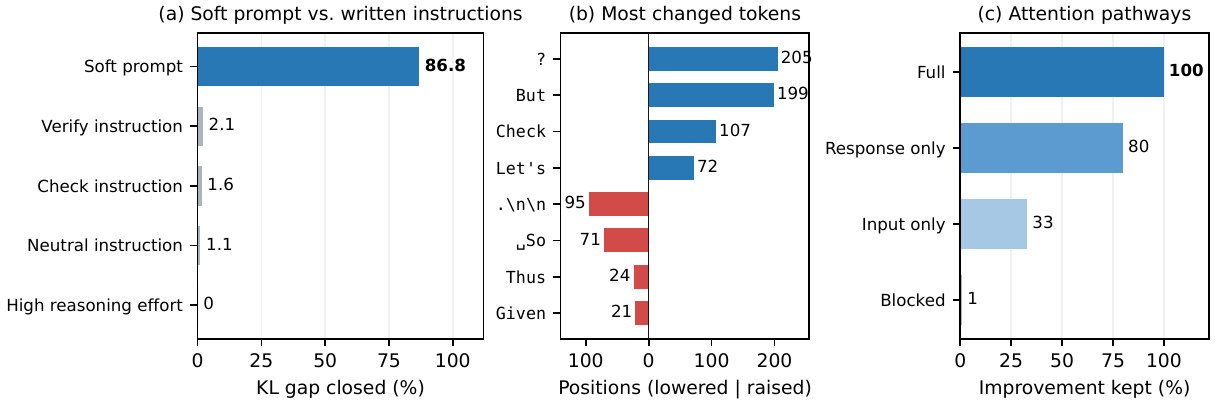}
\caption{What the GPT-OSS math soft prompt does. (a) Share of the teacher--student KL gap closed by the soft prompt, written instructions, and high reasoning effort. (b) Tokens that appear most often among the 2,000 positions where the soft prompt raises (blue) or lowers (red) the probability of the trace token the most. \texttt{\textvisiblespace} marks a leading space. (c) Share of the soft prompt's improvement kept when only some tokens may attend to it. Details in Appendix~\ref{app:interpretability}.}\label{fig:interp-main}
\end{figure*}

\textbf{The soft prompt moves the student most of the way to the teacher, mainly by making it check its intermediate results.} It closes 86.8\% of the teacher--student KL gap. The tokens it raises most are \texttt{?}, \texttt{But}, and \texttt{Check}, and those it lowers most are paragraph breaks, \texttt{So}, and \texttt{Thus} (Figure~\ref{fig:interp-main}b). Instead of stating a result and moving on, the prompted student first questions or checks it, as the teacher does.

\textbf{Written instructions with the same intent do not reproduce the effect.} Instructions asking the model to check and verify its work close at most 2.1\% of the gap, and raising the reasoning effort closes none (Figure~\ref{fig:interp-main}a).

\textbf{The soft prompt acts mainly through response tokens attending to it directly, rather than by changing how the model reads the question.} When only response tokens may attend to the soft prompt, 80\% of its improvement remains, but only 33\% remains when only input tokens may attend to it (Figure~\ref{fig:interp-main}c).

Taken together, a soft prompt does not learn a message that can be put into words. It learns the teacher's reasoning behavior, here checking intermediate results, and applies it through the attention of the response tokens.

%% file: sections/discussion.tex
\section{Discussion}\label{sec:discussion}
\paragraph{Why compose in input space?}
Weight-space consolidation writes every expert into the same parameters, so adding an expert means re-consolidating the model, and the experts can interfere with one another and with the backbone's general capabilities. \cspd{} separates acquiring a capability from combining capabilities. Each prompt is distilled using only its own teacher and data, prompts meet only at inference, and the backbone never changes. This separation lets experts be trained in parallel by different parties, added or replaced without revisiting earlier domains, and removed exactly by dropping their prompt. Our results show that this separation does not cost in-domain performance relative to MOPD, and that it better preserves out-of-domain behavior. The progressive-training comparison further suggests that composition does not require each module to be trained with knowledge of the others.

\paragraph{Plug-in experts versus multi-LoRA.}
Combining LoRA experts from different domains usually takes one of two forms. Merging adds their weight updates with tuned coefficients \citep{ilharco2023arithmetic,yadav2023ties,huang2023lorahub,prabhakar2024loras}, so all experts write into the same weights and can interfere. With the same distillation data, merged LoRA adapters reach 65.50 with TA and 66.19 with TIES, below the 67.83 of composed soft prompts (Table~\ref{tab:compose}). Mixture-style methods instead learn a router or gate over the adapters \citep{buehler2024xlora,zadouri2024molora}, which must be retrained whenever an expert is added. A soft prompt is a plug-in module. Adding an expert trains only its prompt, removing one drops its prompt exactly, and several experts act together within a request by concatenation, with no coefficient or router to fit. Different requests in one batch can also use different sets of prompts on the same unchanged backbone, which gives $2.40\times$ the output throughput of rank-4 LoRA with 32 distinct adapters (Section~\ref{sec:deploy}).

\paragraph{Limitations.}
Our composition study uses one backbone and four domains. Whether composition continues to improve as the number of experts grows, and how it depends on the relations between domains, remain open. Experts do interact, both positively and negatively (Section~\ref{sec:incremental}). Several reported numbers, including single-expert checkpoints and TA coefficients, are selected on the evaluation benchmarks, and small teacher gains make some recovery ratios noisy, so we report absolute scores throughout (Appendix~\ref{app:inventory}). Each prompt adds 256 positions to the context, so attention and KV-cache costs grow with the number of composed experts. Finally, prompts are tied to their backbone and require RL teachers for supervision, so moving to a new backbone requires re-distillation.

\paragraph{Toward continual post-training.}
\cspd{} suggests a different shape for post-training, with a frozen backbone and a growing library of compact expert modules, each distilled when a new capability becomes available and composed as needed. Four directions follow. The first is \emph{scale}, with more backbones, larger prompt libraries, and more arrival orders. The second is \emph{prompt selection}, learning which prompts each request needs instead of choosing them by hand. The third is \emph{mechanism}, understanding why prompts transfer across domains and how several prompts share attention. The fourth is \emph{cost}, with shorter prompts for large libraries and a full training-cost comparison with MOPD and LoRA.

%% file: sections/related_work.tex
\section{Related Work}\label{sec:related}
\noindent\textbf{Capacity requirements of post-training.}
LoRA can match full fine-tuning in selected post-training regimes \citep{schulman2025lora}. Single-layer training, sparse subnetworks, and low-dimensional adapters can also recover substantial RL gains \citep{zhang2026onelayer,mukherjee2025subnetworks,morris2026tiny}, while SGD-based RL exhibits sparse updates \citep{mukherjee2026sgd}. These findings motivate studying input-space representations of expert capabilities.

\noindent\textbf{Distilling post-training capabilities.}
Sequence-level distillation transfers teacher-generated responses, including reasoning capabilities \citep{kim2016,deepseek2025r1}, whereas on-policy methods supervise student-generated trajectories \citep{agarwal2024gkd,gu2024minillm,qwen2025,lu2025opd}. Multi-teacher on-policy distillation consolidates several experts into one student \citep{m2rl2026,ma2026mopd}. Approaching teacher performance in-domain can nevertheless accompany out-of-domain losses \citep{wang2026opdgeneralization}, motivating evaluation beyond the distilled capabilities.

\noindent\textbf{Representations of adaptation.}
Adaptation can reside in weights, continuous inputs, layerwise prefixes, or hidden-state interventions \citep{hu2021,lester2021,li2021prefix,wu2024reft}. PromptKD and soft-prompt teachers use learned prompts to guide distillation into student weights \citep{kim2024promptkd,ma2026teachers}. DDPT instead places demonstration-induced reasoning in dynamically selected prompts \citep{teng2026ddpt}. Soft prompts can also receive distilled safety behavior \citep{spdsafety2026}. Cartridges train key--value cache prefixes to compress long contexts through self-study \citep{eyuboglu2025cartridges}. Concurrent with our work, Hypertune trains key--value cache prefixes directly with RL or SFT, adding co-residence training so that several prefixes can be loaded together \citep{hyperstition2026hypertune}. \cspd{} instead learns prompts in the input embedding space by distilling existing RL experts, trains each prompt in isolation, and composes them by concatenation. An input-space prompt can also be prefilled once into a key--value prefix for serving.

\noindent\textbf{Composition, continual learning, and efficient serving.}
Model soups, Task Arithmetic, and interference-aware merging combine weight updates \citep{wortsman2022soups,ilharco2023arithmetic,yadav2023ties,yu2024dare}, with mixed results for RL experts \citep{sparsitycurse2026,m2rl2026}. Adapter composition combines updates or learns gates \citep{pfeiffer2021fusion,huang2023lorahub,prabhakar2024loras,buehler2024xlora,zadouri2024molora}, while BAR composes experts through a routed mixture-of-experts architecture \citep{bar2026}. In input space, APT learns composable prompts and ATTEMPT mixes source prompts \citep{bowman2023apt,asai2022attempt}. Progressive Prompts freezes earlier prompts during continual learning \citep{razdaibiedina2023progressive}, and Skill Neologisms composes independently learned skill embeddings \citep{skillneologisms2026}. Deployment work optimizes concurrent adapters \citep{chen2023,sheng2024slora}, prefill-only adaptation \citep{lanpouthakoun2026preft}, and speculative decoding \citep{li2025eagle3,chen2026dflash}. \cspd{} connects these research directions by independently distilling RL experts into prompts on a shared frozen backbone, then evaluating their concatenation, out-of-domain behavior, and serving efficiency.

%% file: sections/conclusion.tex
\section{Conclusion}
We presented \cspd{}, which distills each post-RL expert into an independently trained soft prompt on a frozen pre-RL backbone and composes experts by concatenating their prompts. A single prompt recovers at least 90\% of the teacher's gain in 22 of 23 math, code, and agent settings. Four prompts from different domains, trained without access to each other, compose to match MOPD in-domain, and the average rises with every expert added one at a time, as in continual post-training. The composition also attains the highest mean score on held-out benchmarks among the compared methods. Because the experts live in the input, serving them costs little more than serving the backbone alone. These results suggest input-space modules as a practical foundation for continual post-training.

\section*{Impact Statement}
This work studies modular transfer of post-trained behavior. Compact, attachable modules may make specialized behaviors easier to distribute and serve, including behaviors that require safety evaluation. Freezing the backbone does not guarantee that an attached prompt, or a combination of prompts, preserves safety properties, and our XSTest evaluation covers only refusal behavior on the tested requests. Release decisions should consider model and dataset terms, generated training material, and the behavior of the resulting modules alone and in combination.

%% file: sections/appendix.tex
\section{Input Templates and Training Configuration}\label{app:input-templates}
Our multi-expert experiments use the template below with the Qwen3 chat template and thinking enabled. $P$ denotes a 256-vector soft prompt, $s$ the original system-message text (if any), $q$ the target question, and $y$ the assistant response. An empty system turn is added when absent. Each row below is a chat turn, and a newline follows each role header. At inference, generation begins after the assistant header.

\begin{center}
\includegraphics[width=.76\linewidth]{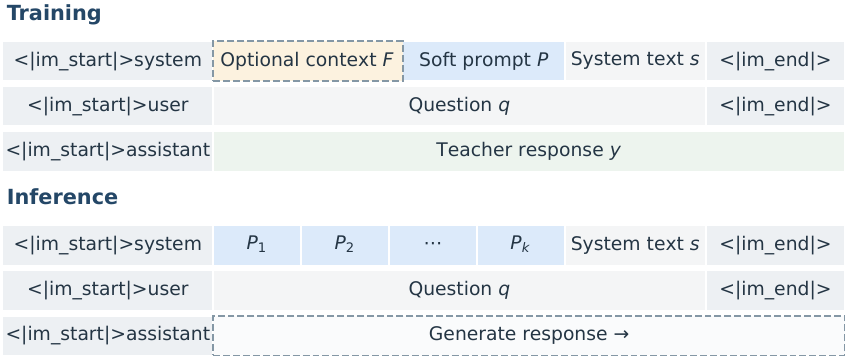}
\end{center}

\paragraph{Prefix-robust training.}
The optional context prefix $F$ is included for half of the examples, selected by a content hash. It contains 128--256 tokens drawn from the prompt-token stream of other examples in the same domain, including chat-template symbols, and is truncated to the selected length, so it can end within a turn. $F$ uses frozen token embeddings and is placed before the SP within the inserted system-prefix region. The original system text, if any, follows the SP. Each training example contains one SP, and only its vectors are optimized. The loss is computed only on the target assistant response, with $F$, SP, system, and user positions masked. $F$ is omitted at inference.

\paragraph{Composition.}
At inference, the SP matrices are concatenated directly, without separators. Their order is left to right as $P_1,\ldots,P_k$, so $P_k$ is the module closest to the user question. Multiple independently trained SPs occur together only at inference.

\paragraph{Training configuration.}
Prompt vectors are initialized from sampled vocabulary embeddings with seed 42 and kept in FP32, while the backbone stays frozen in BF16. The main prompts use Muon with momentum 0.95 and Nesterov momentum. Each update orthogonalizes the gradient with five Newton--Schulz iterations and rescales it to an RMS of 0.2 times the learning rate, following Moonlight \citep{liu2025muon}, so that the learning rate is on the same scale as for AdamW. The learning rate follows a cosine schedule from 0.03 to 0.003 over 200 updates without warmup. Each update uses a global batch of 64 responses, and gradients are clipped to norm 1.0. All four domains use prefix-robust training, and we use the final checkpoint. The AdamW rows in Table~\ref{tab:context-ablation} use $\beta=(0.9,0.999)$, no weight decay, and a cosine schedule from 0.01 to 0.001, so the Muon learning rate is three times the AdamW one. The optimizer and learning rate were screened on IFBench rather than on a separate validation set. The main composition uses MCSI. Continual-addition evaluations prepend modules arriving in I, S, C, M order, producing the final MCSI composition. Single-expert experiments outside the multi-expert setting retain their original per-model protocols.

\section{Additional Single-Expert Results}\label{app:single}
\subsection{Agent settings}\label{app:agents}\label{app:agents-results}
TMax uses Terminal-Bench 2.1 with Harbor and Vanillux2, SkyRL uses OpenHands with CodeActAgent, DeepSWE uses R2E-Gym, and Lego uses Harbor with the OpenHands SDK. Nemotron uses a staged retrieval, localization, and patch-generation pipeline. Its prompted student receives the teacher's localization, whereas the student and teacher baselines localize end to end, so its 92.27\% recovery is conditional on this additional information. The prompts are distilled on \href{https://huggingface.co/datasets/nvidia/Nemotron-Cascade-RL-SWE}{Nemotron-Cascade-RL-SWE} for Nemotron, \href{https://huggingface.co/datasets/NovaSky-AI/SkyRL-v0-80-data}{SkyRL-v0-80} and \href{https://huggingface.co/datasets/NovaSky-AI/SkyRL-v0-220-data}{SkyRL-v0-220} for SkyRL, \href{https://huggingface.co/datasets/R2E-Gym/R2E-Gym-Subset}{R2E-Gym-Subset} for DeepSWE, \href{https://huggingface.co/datasets/Lego-X/Lego-RL-2699}{Lego-RL-2699} for Lego, \href{https://huggingface.co/datasets/allenai/tmax-15k-open-instruct}{tmax-15k-open-instruct} for TMax-9B, and 10,401 teacher episodes for TMax-27B. The teacher--student gaps for SkyRL and TMax-27B are only 1.40 and 2.62 points, so their recovery ratios should be read together with absolute success rates.

\subsection{Training dynamics, prompt length, and supervision}\label{app:train-analysis}
\begin{figure}[!htbp]
\centering
\includegraphics[width=.62\linewidth]{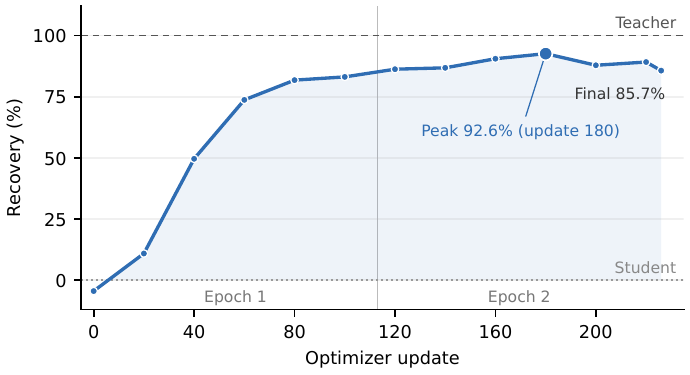}
\caption{GPT-OSS math recovery during training with an earlier, untuned AdamW configuration. All 13 evaluated checkpoints are shown without smoothing, each on 560 questions and 3,920 outputs. The dashed and dotted lines mark the teacher (90.63\%) and the student (77.50\%), and the vertical line marks the end of the first epoch at update 113.}
\label{fig:curve}
\end{figure}

\paragraph{Training dynamics.}
Figure~\ref{fig:curve} shows how recovery evolves during training for the GPT-OSS math prompt. This run comes from an earlier, untuned AdamW configuration (constant learning rate 0.01, two epochs), so its scores differ from Table~\ref{tab:single}, and it is included to show the shape of the recovery curve. Recovery rises quickly in the first 60 updates and then plateaus. It peaks at 92.62\% at update 180 and ends at 85.68\%, so the final checkpoint need not be the best.

\begin{figure}[!htbp]
\centering
\includegraphics[width=.62\linewidth]{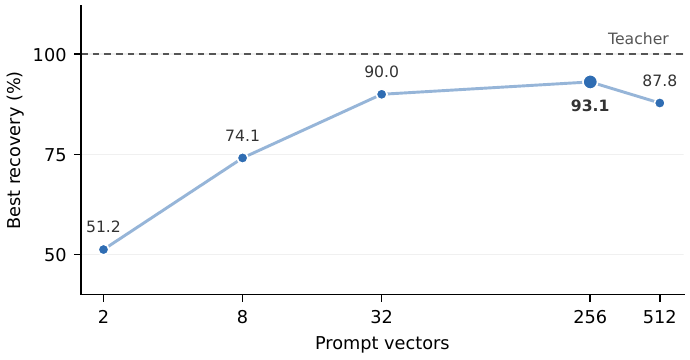}
\caption{GPT-OSS math recovery for five prompt lengths, trained on a smaller set than Figure~\ref{fig:curve}. Each point is the best recovery over checkpoints, and the 256-vector default is enlarged.}
\label{fig:length}
\end{figure}

\paragraph{Prompt length.}
Figure~\ref{fig:length} varies the number of prompt vectors. Recovery is 51.23\%, 74.11\%, and 90.01\% for 2, 8, and 32 vectors, and 93.10\% and 87.83\% for 256 and 512. Most of the gain is reached by 32 vectors. The sweep uses the same AdamW configuration as Figure~\ref{fig:curve} but a smaller training set of 13,804 responses, obtained by capping the responses per question, and its own student and teacher measurements (76.73\% and 90.47\%). Its 256-vector point is therefore a separate run from Figure~\ref{fig:curve}, which explains why its best recovery (93.10\%) differs from the 92.62\% there. Since each vector has $d$ continuous values, prompt length is not an information-theoretic measure of the RL update.

\paragraph{Correct-only versus mixed supervision.}
Table~\ref{tab:supervision} compares two GPT-OSS training sets of 28,946 responses each, one with only correct responses and one that also includes incorrect ones. Both use 256 vectors, AdamW with constant learning rate 0.01, effective batch size 256, and 200 updates, and are evaluated on AIME 2024/2025 avg@16 and MATH500 avg@2. Correct-only supervision gives the highest recovery, 96.06\%. The mixed set, in which only 69\% of responses are correct, is 5.52 points lower but still exceeds 90\%. \cspd{} therefore does not require verifier filtering, and when verified responses are scarce, incorrect responses can be added to enlarge the training set. The two sets keep at most two and four responses per question, so part of this gap may come from question coverage rather than correctness. For reference, a rank-5 LoRA on the K/V projections has a similar parameter count to a 256-vector prompt. With hidden size 2880, the prompt has $256\times2880=737{,}280$ parameters, and the LoRA has $5\times(2880+512)\times2\times24=814{,}080$ over 24 layers with 512-dimensional K/V projections, about 10\% more. It reaches 95.72\% on the correct-only set, close to the prompt.

\begin{table}[!htbp]
\caption{GPT-OSS math distillation with different supervision. Both training sets contain 28,946 responses. Recovery is the best over checkpoints.}\label{tab:supervision}
\centering\small\setlength{\tabcolsep}{5pt}
\begin{tabular}{llrrrrr>{\columncolor[HTML]{E6F1EF}}r}
\toprule
Adapter & Training set & Correct & Incorrect & Correct (\%) & Params & Best update & Recovery (\%)\\
\midrule
SP (256) & Correct-only & 28,946 & 0 & 100.0 & 737,280 & 140 & \textbf{96.06}\\
SP (256) & Mixed & 19,973 & 8,973 & 69.0 & 737,280 & 100 & 90.54\\
LoRA (K/V, $r=5$) & Correct-only & 28,946 & 0 & 100.0 & 814,080 & --- & 95.72\\
\bottomrule
\end{tabular}
\end{table}

\subsection{Evaluation details and limitations}\label{app:inventory}\label{app:supervision}
The GPT-OSS training curve in Figure~\ref{fig:curve} uses 560 questions and 3,920 outputs per checkpoint, including 32 responses per AIME question. MiMo trained on AReaL code responses reaches 52.55\% accuracy and 88.08\% recovery. Mellum2 uses greedy, single-sample evaluation on 1,055 questions and reaches 39.34\% accuracy and 110.34\% recovery. Benchmark windows and sampling counts differ across settings, so comparisons are made within each student--teacher pair.

Where the original RL data are unavailable, we use proxy training data. MiMo math uses DeepScaleR Preview, MiMo code uses AReaL code data, and Mellum2 and Nemotron CodeRL use Klear code data. Single-expert tables report the best checkpoint on the evaluation benchmarks, so they are exploratory rather than held-out estimates. For the AReaL, Klear, Nemotron CodeRL, and OLMo code results, the Klear, MiMo, and Poro math results, and the prompt-length sweep, repeated samples of one question are correlated and are not treated as independent problems. When the teacher--student gap is small, recovery is sensitive to noise, so absolute scores are reported alongside it. Removing a prompt restores the original model exactly, but prompted behavior on unrelated tasks can still change while the prompt is attached.

\section{Prefix-Robust Training, Optimizer, and Concatenation Order}\label{app:context-ablation}
Table~\ref{tab:context-ablation} separates the effects of prefix-robust training and of the optimizer on the multi-expert composition. All rows use seed 42, the final checkpoint after 200 updates, and $K=64$, 8, 16, and 8 samples for AIME24, LCB, GPQA, and IFBench. With AdamW, prefix-robust training raises the average from 64.50 to 66.60 and recovery from 87.0\% to 99.9\%. The gains come from GPQA-D (+3.89), AIME24 (+2.76), and LiveCodeBench (+1.79), while IFBench is unchanged (64.96 versus 65.00). Switching to Muon raises the average further to 67.72 in the same ISCM order, and to 67.83 in MCSI.

\begin{table*}[!htbp]
\caption{Composition ablations in the multi-expert setting. Scores and recovery are percentages. M/C/S/I denote math, code, science, and instruction following.}\label{tab:context-ablation}
\centering\small\setlength{\tabcolsep}{4.5pt}
\begin{tabular}{lrrrrr>{\columncolor[HTML]{E6F1EF}}r}
\toprule
Configuration & AIME24 & LCB v5 & GPQA-D & IFBench & Avg. & Recovery (\%)\\
\midrule
\cspd{} (AdamW, ISCM) w/o prefix-robust training & 76.98 & 64.15 & 51.89 & 65.00 & 64.50 & 87.0\\
\cspd{} (AdamW, ISCM) & 79.74 & 65.94 & 55.78 & 64.96 & 66.60 & 99.9\\
\cspd{} (Muon, ISCM) & 79.90 & 65.72 & 57.92 & 67.33 & 67.72 & 106.8\\
\cspd{} (Muon, MCSI) & 80.47 & 64.67 & 57.77 & 68.42 & \textbf{67.83} & \textbf{107.5}\\
\bottomrule
\end{tabular}
\end{table*}

\begin{figure*}[!htbp]
\centering
\includegraphics[width=\textwidth]{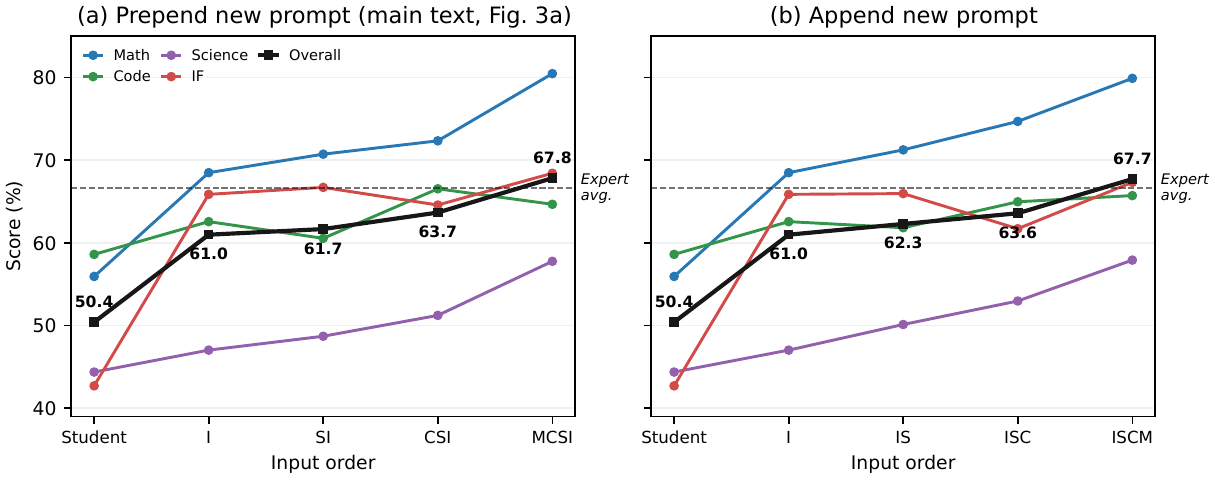}
\caption{Continual addition with experts arriving in the order I, S, C, M. (a) Each new prompt is placed before the existing ones, as in Figure~\ref{fig:incremental}a, ending in MCSI. (b) Each new prompt is placed after the existing ones, ending in ISCM. The dashed line marks the average of the individual experts.}\label{fig:append}
\end{figure*}

\paragraph{Appending new prompts.}\label{app:prepend}
Section~\ref{sec:incremental} places each new prompt before the existing ones. Figure~\ref{fig:append}b instead appends each new prompt after the existing ones, with experts arriving in the same order I, S, C, M, and Figure~\ref{fig:append}a repeats the prepending curve for comparison. The average rises with every addition and reaches 67.72 for the final ISCM composition, close to the 67.83 of MCSI.

\begin{figure*}[!htbp]
\centering
\includegraphics[width=\textwidth]{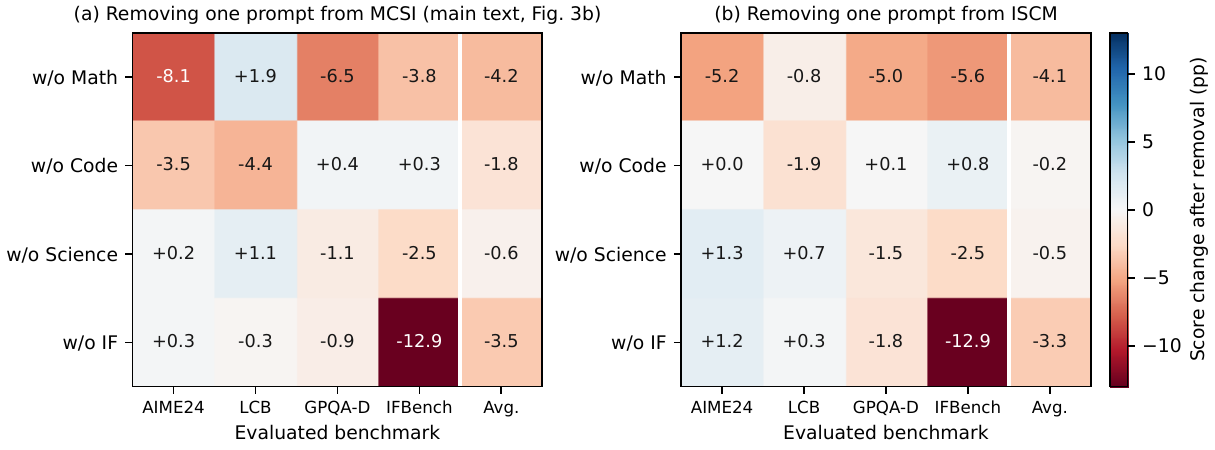}
\caption{Score change on each benchmark (columns) after removing one prompt (rows), preserving the order of the remaining prompts. (a) MCSI, as in Figure~\ref{fig:incremental}b. (b) ISCM. Negative values mean the removed prompt was helping. Both panels share the color scale.}\label{fig:loo-iscm}
\end{figure*}

\paragraph{Removing one prompt from ISCM.}\label{app:loo-iscm}
Figure~\ref{fig:loo-iscm} repeats the leave-one-out analysis of Figure~\ref{fig:incremental}b for the reverse order ISCM, keeping the order of the remaining prompts. The pattern matches MCSI. Removing math or instruction following causes the largest average drops (4.14 and 3.30 points), with AIME24 falling by 5.21 and IFBench by 12.87. Removing science or code costs only 0.47 and 0.25 points on average. Without the code prompt, IFBench rises by 0.84, consistent with the interference between code and instruction following seen in MCSI. Without the science prompt, GPQA-Diamond falls by only 1.45, since the math prompt already contributes much of the science gain.

\section{Baselines and Task Arithmetic Sweep}\label{app:ta-sweep}
\paragraph{Baseline implementation.}
\emph{Expert merge} applies Task Arithmetic (TA) \citep{ilharco2023arithmetic} or TIES \citep{yadav2023ties} directly to the four RL experts. \emph{MOPD} distills multiple experts into the student's weights with multi-teacher on-policy distillation \citep{m2rl2026}. To control for distillation data, we also train one LoRA adapter per domain on the same data used for the prompts and combine the adapters with TA or TIES. LoRA adapters train for 200 updates with the same data and batch order as the prompts, using rank 16, $\alpha=32$, and learning rate $2\times10^{-4}$. TIES uses density 0.2 and scaling 1.0, with task vectors measured relative to the student.

\paragraph{TA coefficient.}
For the four domain task vectors, TA uses $\theta_{\mathrm{TA}}=\theta_0+\lambda\sum_{j=1}^{4}(\theta_j-\theta_0)$, so $\lambda=0.25$ is the arithmetic average of the four domain models. We sweep $\lambda\in\{0.25,0.5,1.0\}$ and report the coefficient with the highest four-domain average, $\lambda=1.0$ for the full-model experts and $\lambda=0.5$ for LoRA (Table~\ref{tab:ta-sweep}). The coefficient is selected on the evaluation benchmarks, which favors the TA baselines.

\begin{table*}[!htbp]
\caption{TA coefficient sweep and TIES results for the full-model and LoRA baselines. Scores and recovery are percentages. Bold marks the selected TA coefficient.}\label{tab:ta-sweep}
\centering\small\setlength{\tabcolsep}{4pt}
\begin{tabular}{llrrrrr>{\columncolor[HTML]{E6F1EF}}r}
\toprule
Model & Merge & AIME24 & LCB v5 & GPQA-D & IFBench & Avg. & Recovery (\%)\\
\midrule
Full model & TA $\lambda=0.25$ & 73.28 & 60.78 & 53.06 & 50.25 & 59.34 & 55.1\\
 & TA $\lambda=0.5$ & 81.04 & 64.15 & 55.84 & 62.00 & 65.76 & 94.7\\
 & TA $\lambda=1.0$ & 78.85 & 64.52 & 54.14 & 71.58 & \textbf{67.27} & \textbf{104.1}\\
 & TIES & 81.25 & 65.42 & 57.07 & 66.12 & 67.47 & 105.2\\
\midrule
LoRA & TA $\lambda=0.25$ & 76.25 & 63.25 & 52.02 & 52.21 & 60.93 & 64.9\\
 & TA $\lambda=0.5$ & 80.21 & 63.10 & 56.22 & 62.46 & \textbf{65.50} & \textbf{93.1}\\
 & TA $\lambda=1.0$ & 78.39 & 56.06 & 52.53 & 69.88 & 64.22 & 85.2\\
 & TIES & 81.41 & 64.00 & 56.66 & 62.67 & 66.19 & 97.3\\
\bottomrule
\end{tabular}
\end{table*}

Increasing $\lambda$ from 0.5 to 1.0 trades off domains. For LoRA, IFBench rises from 62.46 to 69.88, but LiveCodeBench falls from 63.10 to 56.06, and AIME24 and GPQA-D also decline, lowering the average from 65.50 to 64.22. For the full-model experts, IFBench rises from 62.00 to 71.58 while AIME24 and GPQA-D decrease, and the average increases from 65.76 to 67.27. TIES has a higher average than every TA coefficient for both baselines.

\section{Out-of-Domain Evaluation Setting}\label{app:ood-protocol}
BFCL uses 15 categories from BFCL v3 \citep{patil2025bfcl}, 4,041 questions with one sample each, excluding the missing-function and missing-parameter multi-turn categories because they are slow to evaluate. Multi-turn categories contribute one third of the aggregate. TruthfulQA \citep{lin2022truthfulqa} and XSTest \citep{rottger2024xstest} use four samples per question. For XSTest, Qwen3Guard judges whether safe requests are answered and unsafe requests refused, and we report this judgment-based accuracy rather than string matching. The six-benchmark mean weights TriviaQA, BBEH, FinQA, BFCL, TruthfulQA, and XSTest equally.

\section{Serving Memory and Computation}\label{app:serving}
A soft prompt adds KV positions in every attention layer, which generated tokens attend to. For a standard attention stack, the additional KV memory per request is approximately
\begin{equation}
M_{\mathrm{KV}}\simeq 2LK h_{\mathrm{KV}}d_h b,
\end{equation}
where $L$ is the number of layers, $K$ the number of prompt vectors, $h_{\mathrm{KV}}$ the number of KV heads, $d_h$ the head dimension, and $b$ the bytes per value. Hybrid and sliding-window architectures need their own accounting. In our serving setting, 256 prompt vectors lengthen a 20,480-token input by 1.25\%. Unmerged multi-LoRA instead applies request-specific low-rank updates at every adapted projection \citep{hu2021}, which systems such as Punica \citep{chen2023} batch efficiently. A single merged LoRA is a different deployment regime and is not compared here.

\section{Soft-Prompt Mechanism Analysis}\label{app:interpretability}\label{sec:interpretability}
This appendix asks what a soft prompt learns and how it changes the frozen student. We find three things. The learned vectors cannot be read as words (Section~\ref{app:interp-geom}). The soft prompt moves the student's next-token distribution most of the way to the teacher's, mainly by making it check intermediate results, and written instructions with the same intent do not reproduce this (Section~\ref{app:interp-dist}). The soft prompt acts mainly through response tokens attending to it directly, rather than by changing how the model reads the question (Section~\ref{app:interp-path}).

\subsection{Setup}\label{app:interp-setup}
Sections~\ref{app:interp-dist} and~\ref{app:interp-path} study one soft prompt, the 256-vector math prompt of GPT-OSS-Swallow-20B-SFT-v0.1 at update 180, whose teacher is GPT-OSS-Swallow-20B-RL-v0.1 \citep{swallow2026gpt}. As in training, the soft prompt sits at the start of the system message. We use 100 held-out questions, all 30 from each of AIME 2024 and AIME 2025 \citep{aime} and eight from each of the five MATH-500 difficulty levels \citep{hendrycks2021math,lightman2024verify}. The teacher, the student, and the student with the soft prompt (the prompted student) each answer every question twice (temperature 0.6, top-$p$ 0.95, top-$k$ 20, up to 32,768 tokens), giving 600 traces. We then compare models without generating new text. Every model reads the same trace, and at each response position we compare its next-token distribution with the teacher's. $D_{\mathrm{KL}}(T\Vert A)$ denotes the KL divergence from the teacher $T$ to a model $A$, averaged over response positions. It is 0 when $A$ predicts exactly like the teacher.

\subsection{Embedding geometry}\label{app:interp-geom}
\textbf{The nearest word of a soft prompt vector does not tell what the vector learned.} A natural way to read a soft prompt is to map each vector to its nearest vocabulary embedding. We test this on the soft prompts of six models (GPT-OSS, Poro, MiMo, Nemotron, Mellum, and AReaL-8B), using only their weights. In every model, 84--100\% of the vectors still have their initialization token as the nearest word. Yet in five of the models the vectors have moved far from that token, with a median cosine similarity of only 0.25--0.34. The nearest word thus stays the same even when the vector changes a lot. GPT-OSS is the exception, as its vectors barely change direction (cosine 0.999). Across models, the changes from initialization lie in fewer directions than in shuffled or Gaussian controls, with effective ranks of 44--157 out of 256. These numbers describe how the vectors moved, not what the change does, which the next two sections test.

\subsection{Output distribution and natural-language instructions}\label{app:interp-dist}
\begin{figure*}[!htbp]
\centering
\includegraphics[width=\textwidth]{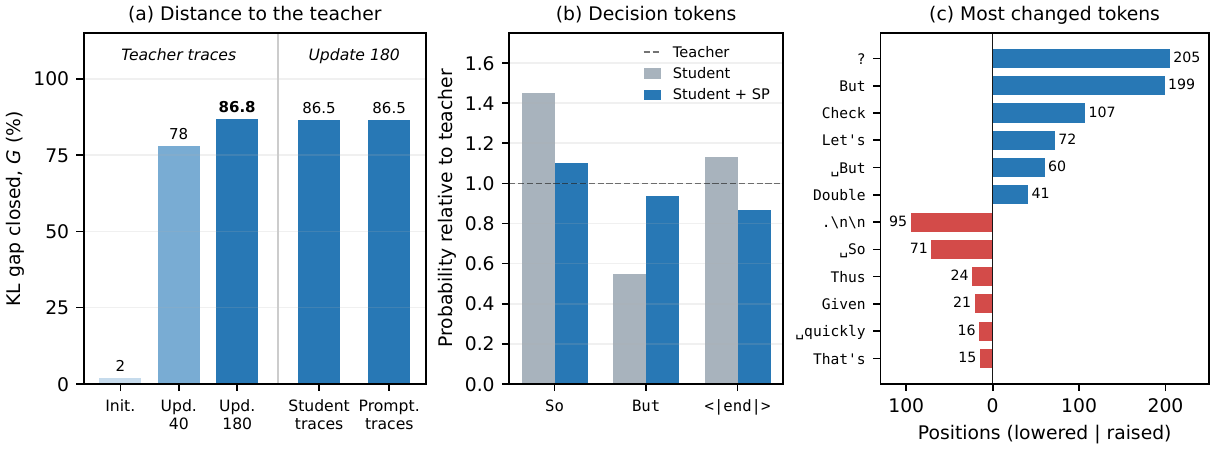}
\caption{Effect of the GPT-OSS math soft prompt on the student's next-token distribution. (a) Share of the student--teacher KL gap closed, $G$ (Eq.~\ref{eq:kl-reduction}), for the soft prompt's initialization, update 40, and update 180 on teacher traces, and for update 180 on student and prompted-student traces. (b) Mean probability of three tokens in the reasoning channel, divided by the teacher's. (c) Tokens that appear most often among the 2,000 positions where the soft prompt raises (blue) or lowers (red) the probability of the trace token the most. \texttt{\textvisiblespace} marks a leading space.}\label{fig:sp-dist}
\end{figure*}

\textbf{The soft prompt moves the student most of the way to the teacher, mainly by making it check its intermediate results.} We score a model $A$ by the share of the student--teacher gap it closes,
\begin{equation}
G(A)=100\left(1-\frac{D_{\mathrm{KL}}(T\Vert A)}{D_{\mathrm{KL}}(T\Vert S)}\right),\label{eq:kl-reduction}
\end{equation}
which is 0 for the student $S$ and 100 for a model that matches the teacher. On the 200 teacher traces, the soft prompt lowers the KL from 0.225 to 0.030, so $G=86.8\%$ (Figure~\ref{fig:sp-dist}a). The soft prompt at update 40 already reaches 78\%, while its random initialization gives 2\%. $G$ is also 86.5\% on the student's traces and on the prompted student's traces, so the effect does not depend on who wrote the text.

The soft prompt makes the student more likely to write \texttt{But}, \texttt{Check}, and \texttt{?} and less likely to write \texttt{So} and \texttt{Thus}, bringing these probabilities close to the teacher's (Figure~\ref{fig:sp-dist}b,c). Without the soft prompt, the student tends to state a result and move on (``Thus\ldots''). With the soft prompt, like the teacher, it first questions or checks the result (``But\ldots'', ``Check\ldots'').

\textbf{Written instructions with the same intent do not reproduce the effect.} If the soft prompt only meant ``check your work'', an instruction saying so should work as well. We place three instructions at the soft prompt's position. The check instruction is ``Do not conclude too quickly. After each intermediate result, question it and check it before moving on.'' The verify instruction is a longer version that also asks the model to double-check results with another method and consider edge cases. The neutral instruction asks the model to read the problem carefully and answer. We also raise the template's reasoning effort from medium to high. These reach $G$ of only 2.1\%, 1.6\%, 1.1\%, and 0.0\%, against 86.8\% for the soft prompt (Figure~\ref{fig:sp-mechanism}a).

\subsection{Attention pathways}\label{app:interp-path}
\textbf{The soft prompt acts mainly through response tokens attending to it directly, rather than by changing how the model reads the question.} The soft prompt can affect the response in two ways. Response tokens can attend to the soft prompt directly, or question tokens can attend to it and the response then attends to the question. We separate the two by masking attention to the soft prompt. The response-only mask lets only response tokens see the soft prompt, the input-only mask lets only the input tokens see it, and the blocked mask hides it from all tokens. All masks keep the model's built-in attention sinks. We use 40 teacher traces and 20 prompted-student traces, each cut to its first 4,096 response tokens, and report the share of the soft prompt's improvement that a masked model $A$ keeps,
\begin{equation}
K(A)=100\,\frac{D_{\mathrm{KL}}(T\Vert S)-D_{\mathrm{KL}}(T\Vert A)}{D_{\mathrm{KL}}(T\Vert S)-D_{\mathrm{KL}}(T\Vert S{+}P)},\label{eq:kept}
\end{equation}
where $S{+}P$ is the student with the unmasked soft prompt. On these traces the two KL values in the denominator are 0.213 and 0.032.

\begin{figure*}[!htbp]
\centering
\includegraphics[width=\textwidth]{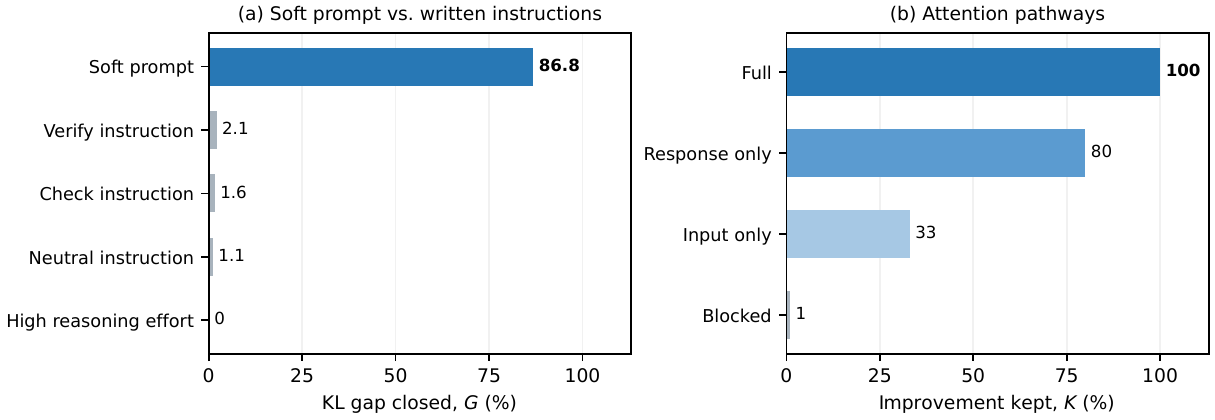}
\caption{Instruction and attention analyses of the GPT-OSS math soft prompt. (a) Share of the student--teacher KL gap closed, $G$ (Eq.~\ref{eq:kl-reduction}), on 200 teacher traces for the soft prompt, two checking instructions, a neutral instruction, and high reasoning effort. (b) Share of the soft prompt's improvement kept, $K$ (Eq.~\ref{eq:kept}), under attention masks on 60 traces. Response only lets only response tokens see the soft prompt, input only lets only input tokens see it, and blocked allows neither. Neither panel reports task scores.}\label{fig:sp-mechanism}
\end{figure*}

Response-only keeps 80\%, input-only keeps 33\%, and blocked keeps about 1\% (Figure~\ref{fig:sp-mechanism}b). Most of the effect therefore comes from response tokens reading the soft prompt. This reading happens mainly in the early layers with full attention (GPT-OSS alternates them with sliding-window layers). Different response tokens also take different information from the soft prompt. At layer 5 under the response-only mask, giving every response token the same average of what it reads from the soft prompt lowers the kept share from 80\% to 28\%. These results come from a single soft prompt without confidence intervals, so finer comparisons between pathways are tentative, and how composed soft prompts act remains to be tested.

